\documentclass{article}

\usepackage{microtype}
\usepackage{graphicx}
\usepackage{subfigure}
\usepackage{booktabs} 

\usepackage{hyperref}

\usepackage{cite}
\usepackage{amsmath,amssymb,amsfonts}
\usepackage{graphicx}
\usepackage{textcomp}
\usepackage{xcolor}
\usepackage{fancyhdr}
\usepackage{multicol}
\usepackage{multirow}
\usepackage[defaultlines=2,all]{nowidow}

\usepackage{soul}
\usepackage[utf8]{inputenc}
\usepackage{graphicx}
\usepackage{booktabs}

\usepackage{multicol}

\usepackage{graphicx}
\usepackage{booktabs} 

\usepackage[ruled,noend]{algorithm2e}

\SetCommentSty{mycommfont}

\usepackage{here}

\usepackage{amsmath,amssymb,amsfonts,amsbsy,amsfonts,latexsym}
\usepackage{multirow}
\usepackage{makecell}
\usepackage[labelfont=bf,textfont=it,belowskip=0pt,aboveskip=5pt,tableposition=top]{caption}
\usepackage{xcolor}
\usepackage{colortbl}

\definecolor{colorA}{RGB}{189,201,225}
\definecolor{colorB}{RGB}{103,169,207}
\definecolor{colorC}{RGB}{ 28,144,153}
\definecolor{colorD}{RGB}{  1,108, 89}

\newcolumntype{R}{>{\columncolor{gray!40}}r}
\newcolumntype{L}{>{\columncolor{gray!40}}l}
\newcolumntype{C}{>{\columncolor{gray!40}}c}

\usepackage{tabularx,colortbl,xcolor}
\usepackage{multirow}
\usepackage[normalem]{ulem}
\useunder{\uline}{\ul}{}

\usepackage{enumitem}

\usepackage{xparse}

\DeclareGraphicsExtensions{.pdf,.png}

\SetKwInput{KwInput}{Input}

\usepackage{longtable}
\usepackage{pgfplots}
\usepackage{outlines}

\newcommand\hb{ \rowcolor{teal!7}}
\newcommand\hc{ \rowcolor{teal!15}}
\newcommand\hd{ \rowcolor{teal!20}}

\newcommand\blfootnote[1]{%
  \begingroup
  \renewcommand\thefootnote{}\footnote{#1}%
  \addtocounter{footnote}{-1}%
  \endgroup
}

\usepackage[accepted]{icml2024}

\usepackage{amsmath}
\usepackage{amssymb}
\usepackage{mathtools}
\usepackage{amsthm}

\usepackage[capitalize,noabbrev]{cleveref}

\theoremstyle{plain}

\theoremstyle{definition}

\theoremstyle{remark}

\usepackage{subcaption}
\usepackage[textsize=tiny]{todonotes}
\usepackage{placeins}

\usepackage{amsmath}

\DeclareMathOperator*{\argmin}{arg\,min}

\newcommand{\OURS}{{SqueezeLLM}\xspace}

\begin{document}

\twocolumn[
\icmltitle{\OURS: Dense-and-Sparse Quantization}



\icmlsetsymbol{equal}{*}

\begin{icmlauthorlist}
\icmlauthor{Sehoon Kim}{equal,berkeley}
\icmlauthor{Coleman Hooper}{equal,berkeley}
\icmlauthor{Amir Gholami}{equal,berkeley,icsi}
\icmlauthor{Zhen Dong}{berkeley}
\icmlauthor{Xiuyu Li}{berkeley}
\icmlauthor{Sheng Shen}{berkeley}
\icmlauthor{Michael W. Mahoney}{berkeley,icsi,lbnl}
\icmlauthor{Kurt Keutzer}{berkeley}

\end{icmlauthorlist}

\icmlaffiliation{berkeley}{UC Berkeley}
\icmlaffiliation{icsi}{ICSI}
\icmlaffiliation{lbnl}{LBNL}
\icmlcorrespondingauthor{Amir Gholami}{amirgh@berkeley.edu}
\icmlkeywords{Machine Learning, ICML}

\vskip 0.3in
]



\printAffiliationsAndNotice{\icmlEqualContribution} 

\begin{abstract}
Generative Large Language Models (LLMs) have demonstrated remarkable results for a wide range of tasks.
However, deploying these models for inference has been a significant challenge due to their unprecedented resource requirements.
This has forced existing deployment frameworks to use multi-GPU inference pipelines, which are often complex and costly, or to use smaller and less performant models.
In this work, we demonstrate that the main bottleneck for generative inference with LLMs is memory bandwidth, rather than compute, specifically for single batch inference.
While quantization has emerged as a promising solution by representing  weights with reduced precision, previous efforts have often resulted in notable performance degradation.
To address this, we introduce \OURS, a post-training quantization framework that not only enables lossless compression to ultra-low precisions of up to 3-bit, but also achieves higher quantization performance under the same memory constraint. 
Our framework incorporates two novel ideas:
(i) \emph{sensitivity-based non-uniform quantization}, which searches for the optimal bit precision assignment based on second-order information; and
(ii) the \emph{Dense-and-Sparse decomposition} that stores outliers and sensitive weight values in an efficient sparse format.
When applied to the LLaMA models, our 3-bit quantization
significantly reduces the perplexity gap from the FP16 baseline by up to 2.1$\times$ as compared to the state-of-the-art methods with the same memory requirement.
Furthermore, when deployed on an A6000 GPU, our quantized models achieve up to 2.3$\times$ speedup compared to the baseline.
Our code is available at \href{https://github.com/SqueezeAILab/SqueezeLLM}{https://github.com/SqueezeAILab/SqueezeLLM}.
\end{abstract}
\section{Introduction}

\begin{figure*}[t!]
\centering{
\centerline{
  \includegraphics[width=0.85\textwidth]{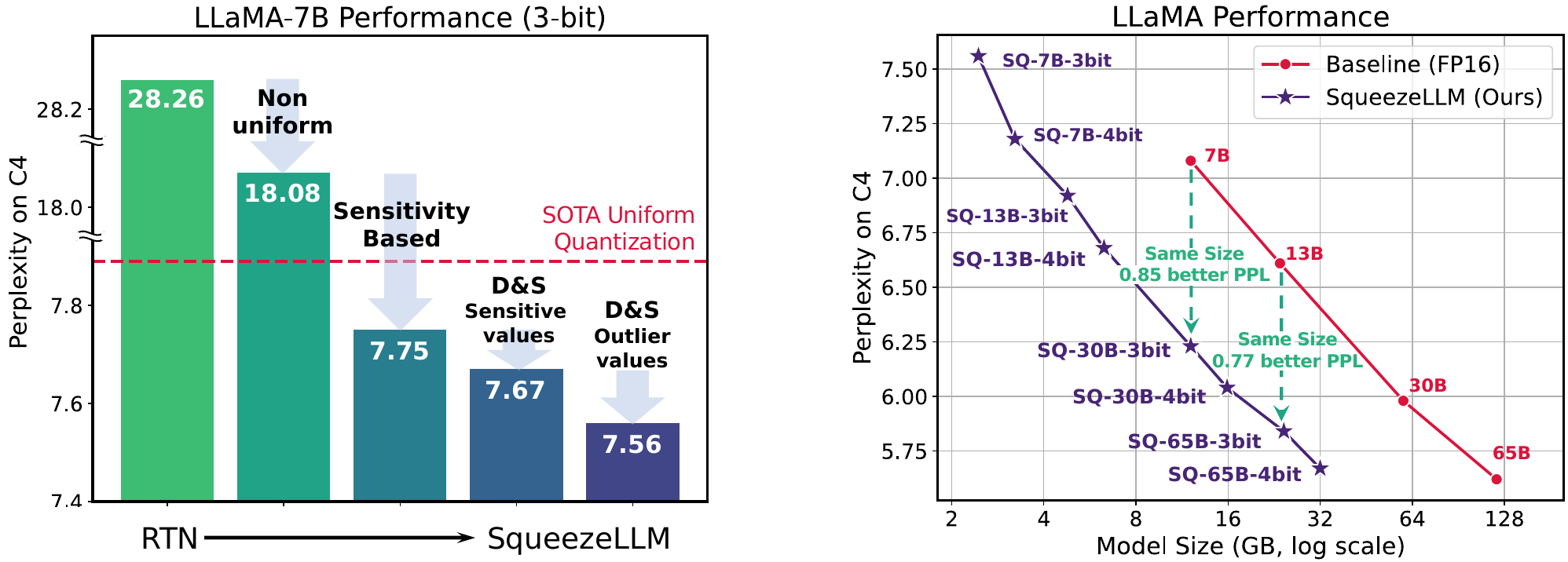}
  }
  \vspace{-1mm}
  \caption{ 
  (Left) 
 \OURS incorporates two key approaches: (i) sensitivity-based non-uniform quantization (Sec.~\ref{subsec:nonuniform}), where quantization bins are allocated closer to sensitive values, and (ii) the Dense-and-Sparse decomposition (Sec.~\ref{subsec:dense_and_sparse}), which retains both sensitive values and outlier values as full-precision sparse format.
 When applied to LLaMA-7B with 3-bit quantization, our method outperforms the state-of-the-art methods~\cite{frantar2022gptq,lin2023awq} by a large perplexity margin of over 0.3 on the C4 benchmark.
  (Right) By applying our methods to LLaMA models of varying sizes, we can achieve improved trade-offs between perplexity and model size.
}
\vspace{-5mm}
  \label{fig:thumbnail}
  }
\end{figure*}
Recent advances in Large Language Models (LLMs) trained on massive text corpora, with up to hundreds of billions of parameters, have showcased their remarkable problem-solving capabilities across various domains~\cite{brown2020language, raffel2020exploring, scao2022bloom, du2022glam, hoffmann2022training,chowdhery2023palm, smith2022using, zhang2022opt, thoppilan2022lamda, touvron2023llama}.
However, deploying these models for inference has been a significant challenge due to their demanding resource requirements.
For instance, the LLaMA-65B model requires at least 130GB of RAM to deploy in FP16, which exceeds current GPU capacity. 
Even storing such large-sized models has become costly and complex.

As will be discussed in Sec.~\ref{sec:memorywall},
the main performance bottleneck in LLM inference for generative tasks is memory bandwidth rather than compute. 
 This means that the speed at which we can load and store parameters becomes the primary latency bottleneck for memory-bound problems, rather than arithmetic computations. 
 However, recent advancements in memory bandwidth technology have been significantly slow, compared to the improvements in computes, leading to the phenomenon known as the Memory Wall \cite{patterson2004latency,aimemory}.
  Consequently, researchers have turned their attention to exploring algorithmic methods to overcome this challenge. 
  
  One promising approach is quantization, where model parameters are stored at lower precision, instead of the typical 16 or 32-bit precision used for training. 
  For instance, it has been demonstrated that LLM models can be stored in 8-bit precision without performance degradation \cite{yao2022zeroquant}, where 8-bit quantization not only improves the storage requirements by half but also has the potential to improve inference latency and throughput.
As a result, there has been significant research interest in quantizing models to even lower precisions. 
A pioneering approach is GPTQ~\cite{frantar2022gptq} which uses a training-free quantization technique that achieves near-lossless 4-bit quantization for LLM models with over tens of billions of parameters.
However, achieving high quantization performance remains challenging, particularly with lower bit precision and for relatively smaller models (e.g., $<50$B parameters).

\noindent
\textbf{Contributions.} 
In this paper, we conduct an extensive study of low-bit precision quantization, and we identify limitations in existing approaches. 
Based on the insight that \textit{the memory, rather than the compute, is the primary bottleneck } in LLM inference with generative tasks, we introduce \OURS, a post-training quantization framework with a novel \textit{sensitivity-based non-uniform quantization} and \textit{Dense-and-Sparse decomposition}.
These techniques enable lossless compression even at precisions as low as 3 bits with reduced model sizes and faster inference without compromising model performance. 
Our detailed contributions include:

\vspace{-2mm}
\begin{itemize}[leftmargin=3.3mm]
    
    \item \textbf{Sensitivity-based Non-Uniform Quantization:} 
    We demonstrate that uniform quantization of prior works
    is sub-optimal for LLM inference for two reasons.
    First, the weight distributions in LLMs exhibit clear non-uniform patterns (Fig.~\ref{fig:nonuniform}).
    Second, the inference computation in prior works does not fully benefit from uniform quantization as the arithmetic is performed in FP16 precision, not in reduced precision.
    To address these, we propose a novel sensitivity-based non-uniform quantization method to achieve more optimal LLM quantization, which significantly improves the perplexity of 3-bit LLaMA-7B from 28.26 of uniform quantization to 7.75 on C4 (Sec.~\ref{subsec:nonuniform}).

    \item \textbf{Dense-and-Sparse Quantization:}
    The weights in LLMs contain significant outliers, making low-bit quantization extremely challenging. 
    To address this, we propose a simple solution that decomposes weights into dense and sparse components.
    The sparse part holds outlier values in full precision using efficient sparse storage methods, and the dense part can have a more compact range to aid quantization. 
    By  extracting only 0.45\% of the weight values as the sparse component, we further improve
    the perplexity of LLaMA-7B from 7.75 to 7.58 on C4  (Sec.~\ref{subsec:dense_and_sparse}). 
    
    \item \textbf{Evaluation:} We extensively test \OURS on various models on language modeling tasks using the C4 and WikiText2 datasets as well as on the MMLU~\cite{hendryckstest2021} and Vicuna benchmarks~\cite{vicuna2023} (Sec.~\ref{subsec:instruction-following}).
     Furthermore, our deployed models on A6000 GPUs also exhibit significant latency gains of up to 2.4$\times$ compared to the FP16 baseline,
     showcasing the effectiveness of our method in terms of both quantization performance and inference efficiency (Sec.~\ref{subsec:deployment}).

\end{itemize}

\section{Related Work}

\textbf{LLM Quantization.} In Appendix~\ref{appendix:related}, we offer an overview and related works of Transformer quantization, with an emphasis on Post-Training Quantization (PTQ), which is the primary focus of our work.
With the increasing popularity of LLMs, \textit{weight-only quantization} has surfaced as a promising approach to reduce memory consumption and enhance inference efficiency. GPTQ~\cite{frantar2022gptq} has been a pioneering work, and  AWQ~\cite{lin2023awq} and SpQR~\cite{dettmers2023spqr} have also suggested the weight-only quantization schemes concurrent to our work.
Our work, however, is different in two key aspects.
First, our work employs non-uniform quantization, as opposed to uniform quantization of the aforementioned works.
In particular, our sensitivity-based non-uniform quantization
not only better represents non-uniform distributions of weights, but it also strategically reduces the impact on more sensitive values, thereby enabling more aggressive quantization without performance degradation.
Second, while previous works quantize weights in a way that layer-wise output activations remain unaffected, our approach targets preserving the model's final output. 
This strategy of minimizing the final loss, as shown in Appendix~\ref{subsec:ablation-obd}, leads to better quantization performance since it is a direct measure of the end-to-end
performance degradation after quantization.

\textbf{Non-uniform Quantization.} 
For low-bit LLM quantization, \cite{dettmers2024qlora} has recently introduced the NF datatype, highlighting the importance of non-uniform quantization.
 However, our approach differs by offering a more dynamic non-uniform representation that accounts for both weight distributions and sensitivity of values, as opposed to the static, hard-coded NF datatype that assumes the normal distribution of the weights.
While previous studies~\cite{han2015deep,xu2018deep} have used k-means clustering in quantization, our work pioneers its application in LLM quantization. Furthermore, we introduce the novel sensitivity-based weighted k-means clustering strategy, enabling lossless sub-4-bit quantization by significantly reducing performance degradation, in contrast to the sensitivity-agnostic counterpart (Fig.~\ref{fig:thumbnail}).

\textbf{Outlier-Aware Quantization.} 
Among the various challenges in low-bit Transformer quantization, one key issue is the presence of outliers~\cite{kovaleva2021bert}, which can unnecessarily increase the quantization range.
To address this issue, outlier-aware quantization methods have been investigated~\cite{bondarenko2021understanding, dettmersgpt3, wei2022outlier, wei2023outlier, xiao2023smooth}.
Notably, \cite{dettmersgpt3} keeps outlier activations in floating-point, while \cite{wei2022outlier} transfers outlier factors to later layers without affecting functionality.
These focus on activations, which is not a concern in our work where all activations are in floating-point. Our Dense-and-Sparse quantization instead tackles \textit{weight} outliers for low-bit LLM quantization.

Concurrently to our work, SpQR~\cite{dettmers2023spqr} also explores outlier extraction in the context of weight quantization.
While SpQR has shown a promising result on outlier extraction, \OURS, leveraging sensitivity-based non-uniform quantization, achieves precise quantization with significantly lower (e.g., 0.05\%) or even zero sparsity levels. This is critical for both reducing model size and improving inference speed, as higher sparsity often degrades latency.
Furthermore, \OURS uses outlier extraction as a direct solution to prevent outliers from negatively impacting quantization performance, bypassing the need for using the grouping strategy as an indirect solution. 
 This contrasts with SpQR, which relies on fine-grained grouping that leads to increased model size and a more complex quantization process such as the bi-level quantization scheme.

\textbf{Dense-and-Sparse Decomposition.} 
Matrix decomposition into dense and sparse components has been explored in attention map decomposition~\cite{chen2021scatterbrain,dass2023vitality}, leveraging the fact that attention patterns often present low-rank characteristics with a few outliers. 
To the best of our knowledge, however, our research is the first to apply the dense-and-sparse decomposition strategy to weight matrices to improve quantization performance. 
Additionally, we uniquely incorporate both outlier and sensitive values within the sparse matrix, which yields considerable improvement in post-quantization performance.

\section{Memory Wall}
\label{sec:memorywall}

\begin{figure}[t!]
\centering
\includegraphics[width=1.1\linewidth]{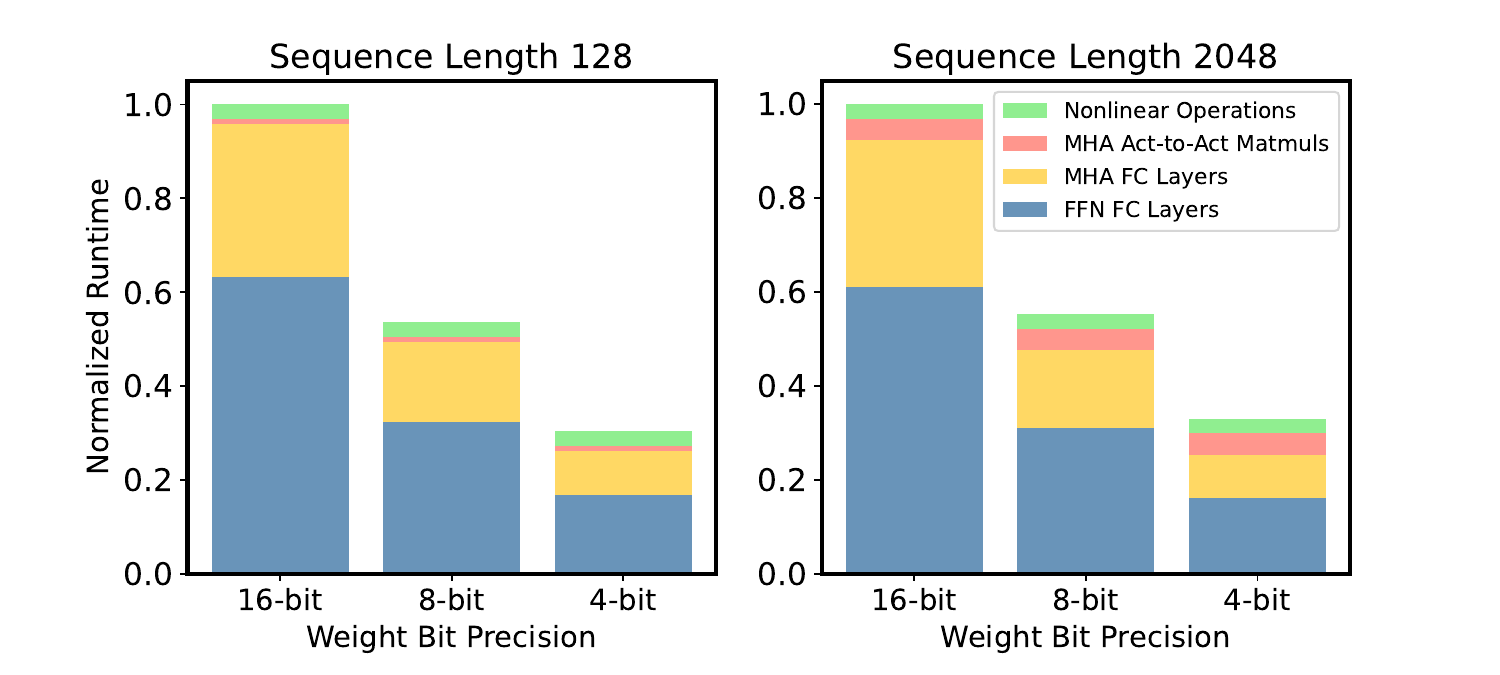}
\vspace*{-7mm}
 \caption{ 
 Normalized runtime for LLaMA-7B when reducing the bit precision for the weights with sequence lengths of 128 (left) and 2048 (right).
Results were obtained using a roofline-based performance model for an A5000 GPU.
Reducing only the precision of the weights (and not the activations) is sufficient to obtain significant latency reductions.
  }
  \vspace{-2mm}
  \label{fig:performance-model}
\end{figure}

Inference behavior broadly falls into two categories: \textit{compute-bound} inference that is limited by computational throughput, and \textit{memory-bound} inference that is bottlenecked by the rate at which data can be fed into the processing cores from memory. 
\textit{Arithmetic intensity}, the ratio of compute to memory operations, is a typical metric used to assess this behavior. 
High and low arithmetic intensity indicates a compute-bound and memory-bound problem, respectively.
For memory-bound problems, the speedup can be achieved by reducing the memory traffic rather than compute since the compute units in hardware are often underutilized waiting to receive data from memory.

Generative LLM inference exhibits extremely low arithmetic intensity compared to other workloads\footnote{{To be precise, we limit this discussion to single batch
inference where the arithmetic involves matrix-vector operations. For large batch inference, compute can become important.}}~\cite{kim2023full}.
This is because it consists almost entirely of matrix-vector operations, which limits the data reuse as each weight load can only process a single vector for a single token, and cannot be amortized across the multiple vectors for different tokens.
This low arithmetic intensity needs to be contrasted with the compute operations on a typical GPU which is orders of magnitude higher than the memory operations.\footnote{{For instance, A5000 GPU has peak computational throughput of 222 TeraFLOPs per second, which is 290$\times$ higher than the peak memory bandwidth of 768 GigaBytes per second.}}
The disparity between compute and memory bandwidth, along with the growing memory requirements of deep learning, has been termed the \textit{Memory Wall} problem \cite{aimemory}.
To further illustrate this problem, we used a simple roofline-based performance modeling approach~\cite{kim2023full} to study LLaMA-7B's runtime on an A5000 GPU with different bit precisions (Fig.~\ref{fig:performance-model}). 
While we assume that all computations are kept at FP16, we see that the latency decreases linearly as we reduce the bit precision, indicating that the main bottleneck is memory, not compute.

In summary, in generative LLM inference, loading weights  into memory is the primary bottleneck, while the cost of dequantization and FP16 computation is relatively small. 
Thus, by quantizing just the weights to lower precision, while leaving the activations in full precision, we can attain significant speedup as well as reduced model size.
Given this insight, the appropriate strategy is to \textit{minimize the memory size even if it may add overhead to arithmetic operations}.
\section{Methodology}
\label{sec:methodology}

\subsection{Sensitivity-Based Non-uniform Quantization}
\label{subsec:nonuniform}

\begin{figure}[t]
\centering
\includegraphics[width=\linewidth]{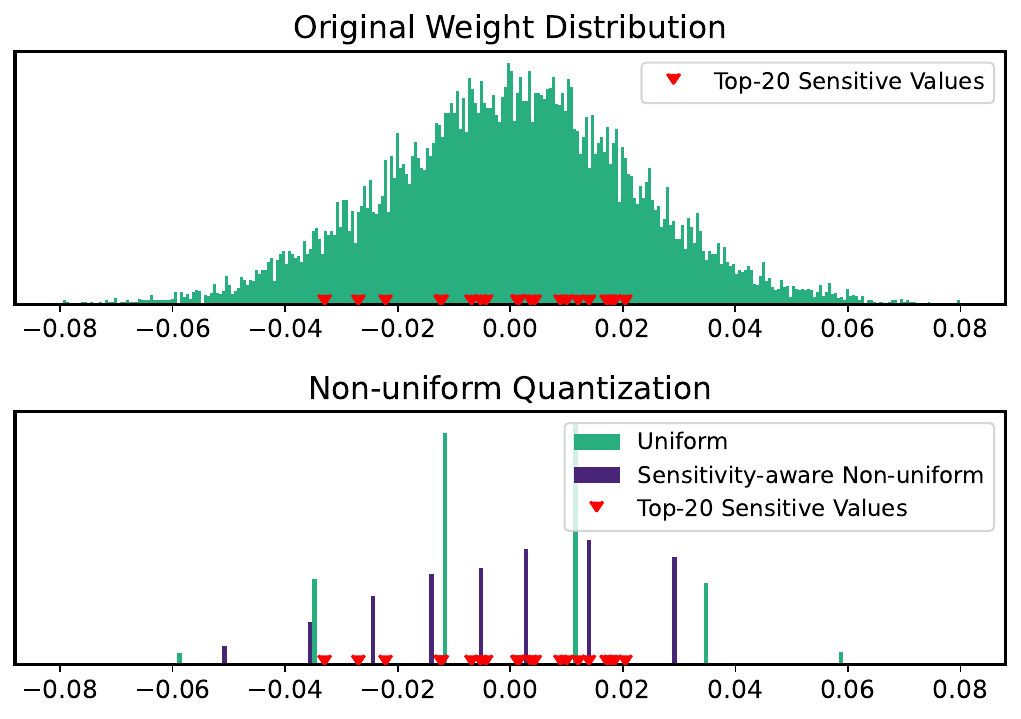}
\vspace*{-8mm}
 \caption{
 (Top) The weight distribution of one output channel in LLaMA-7B.
The top-20 sensitive values are marked in red.
(Bottom) Weight distributions after 3-bit quantization using uniform and sensitivity-based non-uniform quantization. In the latter case, the quantized values are clustered around the sensitive values.
}
  \vspace{-3mm}
  \label{fig:nonuniform}
\end{figure}

As in Fig.~\ref{fig:nonuniform} (Top), weight distributions in LLMs demonstrate non-uniform patterns.
The main task for quantization is to find an optimal way to allocate distinct quantized values (e.g., 8 for 3 bits) in a way that preserves model performance.
A widely used approach in LLM quantization works is uniform quantization, where the weight
range is evenly divided into bins.
This has two main issues.
First, uniformly distributing quantized values is sub-optimal as weight distributions are typically non-uniform.
Second, while the main advantage of uniform quantization is efficient integer computation, this does not lead to end-to-end latency improvement in memory-bound LLM inference.
Therefore, we have chosen non-uniform quantization,
which allows for a more flexible allocation of the representative values.

Finding an optimal non-uniform quantization configuration translates into solving a k-means problem.
Given a weight distribution, the goal is to determine $k$ centroids that best represent the values (e.g., $k$=8 for 3-bit). This optimization problem for non-uniform quantization can be formulated as 
\begin{equation}
\label{eq:l2_kmeans}
Q(w)^* = \argmin_{Q} \lVert W - W_Q \rVert_2^2, 
\end{equation}
\vspace{-3mm}

where $W$ denotes the weights and $W_Q$ is the corresponding quantized weights (i.e., $[Q(w)$ for $w \in W]$), represented by $k$ distinct values $\{q_1, \cdots, q_k\}$.
Here, the optimal solution $Q(w)^*$ can be obtained by  1-dimensional k-means clustering, which clusters the parameters into $k$ clusters and assign the centroid of each cluster as $q_j$'s.
While this already outperforms uniform quantization, we propose an improved \textit{sensitivity-based} clustering algorithm.

\noindent
\textbf{Sensitivity-Based K-means Clustering.}
The quantization objective is to represent the model weights with low-bit precision with minimal perturbation in the model output~\cite{dong2019hawqv2}. 
While quantization introduces perturbations in each layer,
we need to minimize the overall perturbation with respect to the \textit{final loss}, rather than focusing on individual layers, as it provides a more direct measure of the end-to-end performance degradation after quantization~\cite{lecun1990optimal}.
To achieve this, we need to place the k-means centroids near the values that are more sensitive with respect to the final loss, rather than treating all weight values equally, as in Eq.~\ref{eq:l2_kmeans}. 
To determine more sensitive values, we perform Taylor expansion to analyze how the loss changes in response to perturbations in the weights $W$:
\vspace{-1mm}
\begin{align}
\scriptsize \mathcal{L}(W_Q) &\simeq \mathcal{L}(W) - g^\top (W-W_Q) \\
&+ \frac{1}{2}(W-W_Q)^\top H (W-W_Q)
\end{align}
where $g$ and $H = \mathbb{E}[\frac{\partial ^2}{\partial W^2}\mathcal{L}(W)]$ are
the gradient and Hessian of the loss at $W$.
Assuming that the model has converged, the gradient $g$ can be approximated as zero which
gives us the following formula for computing how much the model gets perturbed after quantization:
\begin{equation}
\label{eq:opt2}
    Q(w)^* = \argmin_{Q} (W-W_Q)^\top H (W-W_Q).
\end{equation}
In the new optimization target, as compared to Eq.~\ref{eq:l2_kmeans}, the perturbation of each weight after quantization, i.e., $W - W_Q$, is weighted by the scaling factor introduced by the second-order derivative, $H$.
This highlights the importance of minimizing perturbations for weights with large Hessian values, as they have a greater impact on the overall perturbation of the final output.
In other words, the second-order derivative serves as a measure of importance for each weight value.

\begin{figure*}[t]
\centering
\includegraphics[width=\linewidth]{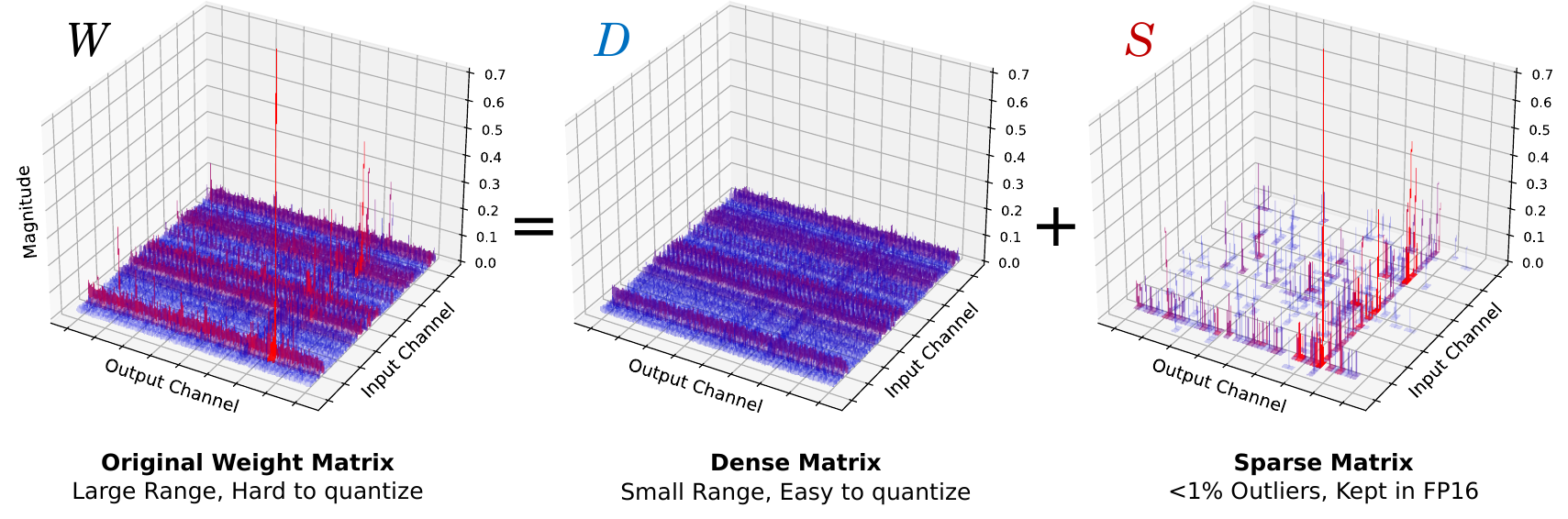}
\vspace*{-3mm}
 \caption{
The illustration of the Dense-and-Sparse decomposition.
The left figure plots the magnitude of a weight matrix ($W$) in the LLaMA 65B model, which contains a few outliers. 
These outliers contribute to the large range of values in the original weight matrix which significantly degrades the quantization performance.
This matrix, however, can be decomposed into a sparse matrix $S$ (Right) that contains the outliers and the remaining dense matrix $D$ (Middle). 
The dense matrix $D$ then exhibits a significantly smaller range, making accurate quantization much easier. 
The sparse matrix $S$ can be kept in full precision with minimal memory and runtime overhead.
  }
  \label{fig:dense-and-sparse}
\end{figure*}

\begin{figure}[t]
\centering
\includegraphics[width=\linewidth]{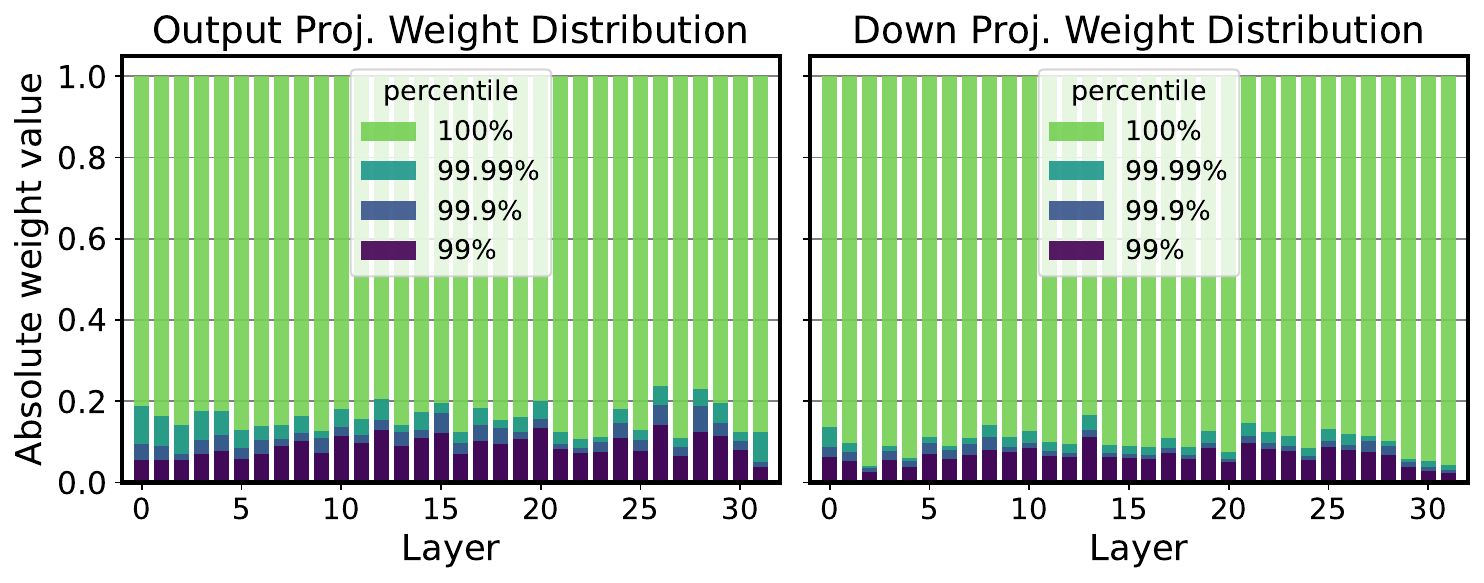}
\vspace*{-6mm}
 \caption{
The distributions of the (normalized) absolute weight values, for the output layers in MHA and the down layers in FFN across different layers in LLaMA-7B.
Note that the distributions exhibit outlier patterns across all layers, with 99\% of the values clustered within $\sim$10\% of the entire range.
  }
  \label{fig:outliers}
\end{figure}

Due to the cost of computing the Hessian, we use an approximation to the Hessian based on the Fisher information matrix $\mathcal{F}$, which can be calculated over a sample dataset $D$ as $H \simeq \mathcal{F} = \frac{1}{|D|}\sum_{d\in D}g_d {g_d}^\top.$
This only requires computing gradient for a set of samples, which can be calculated efficiently with existing frameworks.
To make the optimization objective in Eq.~\ref{eq:opt2} more feasible, 
we further approximate the Fisher information matrix as a diagonal matrix
by assuming that the cross-weight interactions are negligible.
This simplifies our objective target as follows:

\begin{align}
\label{eq:fisher_kmeans}
    Q(w)^* &\simeq \argmin_{Q} (W-W_Q)^\top \mathrm{diag}(\mathcal{F})(W-W_Q) \\
    &= \argmin_{Q} \sum_{i=1}^{N} \mathcal{F}_{ii} \big(w_i-Q(w_i)\big)^2.
\end{align}
An important consequence of Eq.~\ref{eq:fisher_kmeans} is the \textit{weighted} k-means clustering setting, where the centroids will be pulled closer to these sensitive weight values.
In  Fig.~\ref{fig:nonuniform}, we illustrate the top-20 sensitive values based on the Fisher information of the exemplary weight distribution.
At the bottom, the quantized values assigned by uniform quantization (green) are compared to those assigned by the sensitivity-based k-means approach (purple), which achieves a better trade-off by placing centroids near sensitive values, effectively minimizing quantization error.
With 3-bit LLaMA-7B, sensitivity-based non-uniform quantization achieves much lower perplexity of 7.75 compared to the 28.26 perplexity of round-to-nearest uniform quantization on C4 (Fig.~\ref{fig:thumbnail} and Sec.~\ref{subsec:main})

\subsection{Dense-and-Sparse Quantization}
\label{subsec:dense_and_sparse}

Another challenge in low-bit LLM quantization is outlier values~\cite{bondarenko2021understanding, dettmersgpt3, wei2022outlier, wei2023outlier}.
In Fig.~\ref{fig:outliers}, we plot the normalized weight distributions of different layers in LLaMA-7B, which demonstrate that $\sim$99.9\% of the weights are concentrated in a narrow range of $\sim$10\% of the entire distribution.
Naively quantizing the weights with a large range will significantly degrade performance, especially at low precisions.
However, this also implies opportunity as the range of the weight values can be contracted by a factor of 10 simply by removing a small number of outlier values (e.g., 0.1\%), yielding a significant improvement in quantization resolution.
This will then help the sensitivity-based k-means centroids to focus more on the sensitive values rather than a few outliers.

Motivated by this, we introduce a method to filter out outliers from the weight matrix $W$ by performing a simple yet effective decomposition into a sparse matrix ($S$) containing the outliers and the remaining dense matrix ($D$)
that can be quantized much more effectively thanks to its significantly
reduced range of values.
That is, $W = D + S$ where $D = W[T_\mathrm{min} \le w \le T_\mathrm{max}]$ and $S = W[w < T_\mathrm{min} \enspace \mathrm{or} \enspace w > T_\mathrm{max}]$.
Here, $T_\mathrm{min/max}$ are thresholds that define outliers based on the percentile of the distribution.
This Dense-and-Sparse decomposition process is visually illustrated in Fig.~\ref{fig:dense-and-sparse}.

Importantly, the overhead of this decomposition is minimal, since the number of outlier values is small (e.g., $0.5\%$ of the entire values). 
Therefore, the sparse matrix can be stored efficiently using methods like the compressed sparse row (CSR) format.
Inference is also straightforward with the decomposition
as in $WX = DX + SX$, two kernels for dense and sparse multiplication can be overlapped, and the sparse part ($SX$) can benefit from sparse kernels (Sec.~\ref{subsec:dense_and_sparse_kernels}).

\noindent
\textbf{Sensitivity-Based Sparse Matrix.}
In addition to isolating outliers into a sparse matrix, we've also discovered the advantage of precisely representing a small number of highly sensitive weight matrix values.
These values can be easily identified based on the Fisher information 
 (Sec.~\ref{subsec:nonuniform}). 
This not only maintains sensitive values with FP16 to avoid their impact on the model output, 
but also prevents the centroids of Eq. \ref{eq:fisher_kmeans} from skewing towards the sensitive values.
We have observed that extracting only 0.05\% of these sensitive values across layers substantially enhances quantization performance (Appendix~\ref{subsec:ablation}).
Altogether, with 3-bit LLaMA-7B, extracting 0.45\% of outlier and sensitive values further reduces the perplexity from 7.67 to 7.56 (Fig.~\ref{fig:thumbnail} and Sec.~\ref{subsec:main}).

\subsection{Dense-and-Sparse Kernel Implementation}
\label{subsec:dense_and_sparse_kernels}
To efficiently process non-uniformly quantized values,
we implement 3/4-bit CUDA LUT-based kernels for matrix-vector multiplication between compressed weight matrices and uncompressed activation vectors. 
These kernels load the compressed weights and dequantize them piece-by-piece to minimize memory bandwidth utilization. 
The compressed matrices store 3/4-bit indices, which correspond to LUT entries containing FP16 values associated with the bins obtained from non-uniform quantization.
After dequantization, all arithmetic is performed in FP16. 

To optimize the handling of our Dense-and-Sparse representation, we develop kernels for sparse matrix-vector multiplication that load a matrix in CSR format and a dense activation vector, inspired by~\cite{spmv-blog}. 
Since the non-zero entry distributions are highly skewed across rows (Appendix~\ref{subsec:data_skew}), assigning a single thread per row can be inefficient due to uneven workload distribution among threads.
Thus, we implement \textit{balanced hybrid kernels} based on \cite{flegar2017balanced}
by assigning an equal number of nonzeros per thread; this leads to additional synchronization across threads due to rows being processed by multiple threads, but leads to a more balanced work assignment. 
We set the number of threads such that there were 10 nonzeros per thread.
The dense non-uniform kernel and balanced sparse kernels are launched in one call to avoid overhead from summing the outputs from these separate operations.

\vspace{-2mm}
\section{Evaluations}
\label{sec:results}

\vspace{-1mm}
\subsection{Experiment Setup}
\vspace{-1mm}
\label{subsec:setup}
Below is our experiment setup, with more details in Appendix~\ref{appendix:experiment_setup}.

\textbf{Models and Datasets.} 
We have conducted comprehensive evaluations of \OURS using various models including  LLaMA, LLaMA2~\cite{touvron2023llama, touvron2023llama2}, OPT~\cite{zhang2022opt} and Vicuna~\cite{vicuna2023} (v1.1 and v1.3).
We conduct language modeling evaluation using the  C4~\cite{raffel2020exploring} and WikiText2~\cite{merity2016pointer} datasets.
We further evaluate the domain-specific knowledge and problem-solving ability using MMLU~\cite{hendryckstest2021}, and the instruction-following ability using the methodology in~\cite{vicuna2023}.

\vspace{-1mm}
\noindent
\textbf{Baseline Methods.} 
We compare \OURS against various PTQ methods for LLMs including RTN, GPTQ~\cite{frantar2022gptq}, AWQ~\cite{lin2023awq} and SpQR~\cite{dettmers2023spqr}.
To ensure a fair comparison, we use GPTQ \textit{with} activation ordering throughout all experiments unless specified, which addresses the significant performance drop that would otherwise occur.

\vspace{-1mm}
\noindent
\textbf{Quantization Details.}
For \OURS, we adopt channel-wise quantization where each output channel is assigned a separate lookup table. 
We use 2 different sparsity levels: 0\% (dense-only) and 0.45\% (0.05\% sensitive values and 0.4\% outlier values, as discussed in Sec.~\ref{subsec:dense_and_sparse}). 
For measuring sensitivity, we use 100 random samples from the Vicuna training set for Vicuna models and C4 training set for the others. 
While grouping can also be incorporated with our method, we found it sub-optimal as compared
to extracting sensitive/outlier values with sparsity (Appendix~\ref{subsec:ablation-grouping}).

\vspace{-1mm}
\noindent
\textbf{Latency Profiling. }
We measure the latency and peak memory usage for generating 128 and 1024 tokens on an A6000 machine using the Torch CUDA profiler.
As an official implementation of GPTQ (in particular, the grouped version) 
is not available, we implement an optimized kernel for single-batch inference based on the most active open-source codebase~\cite{gptq-for-llama}.

\vspace{-1mm}
\subsection{Main Results}
\label{subsec:main}
\vspace{-1mm}

\begin{table*}[t!]
\caption{\footnotesize Perplexity comparison of LLaMA models quantized into 3 and 4 bits using different methods including RTN, GPTQ, AWQ and SpQR on C4 and WikiText-2.
We compare the performance of different methodologies by grouping them based on their model sizes. 
In the first group, we compare dense-only \OURS with non-grouped GPTQ. 
In the second group, we compare \OURS with a sparsity level of 0.45\% to GPTQ and AWQ with a group size of 128. 
 For comparison, we add speedup and peak memory usage numbers, which we provide more details in Tab.~\ref{tab:llama-7b-main-full}.
Further results for \textbf{LLaMA-30/65B} can be found
in Tab.~\ref{tab:llama-7b-main-full}, and results on other models including \textbf{LLaMA-2 7/13/70B} are provided in Appendix~\ref{appendix:additional-results}.
}
\label{tab:llama-7b-main}
\vspace{-6mm}
\centering{
\footnotesize{
\setlength{\tabcolsep}{7pt}{
    \vspace{2mm}
    
    \begin{subtable}
        \centering
        \scriptsize{

        \vspace{-1mm}
        \begin{tabular}{c|c|cc|cc|c|cc|cc}
         \toprule
            {\textbf{LLaMA-7B}} & \multicolumn{5}{c|}{\textbf{3-bit}}  & \multicolumn{5}{c}{\textbf{4-bit}} \\
            \midrule
           \multirow{2}{*}{\textbf{Method}} & \textbf{Avg. Bits} & \multicolumn{2}{c|}{\textbf{PPL} (↓)} & {\textbf{Speedup}} & {\textbf{Mem.} }  & \textbf{Avg. Bits} & \multicolumn{2}{c|}{\textbf{PPL} (↓)} & {\textbf{Speedup}} & {\textbf{Mem.} } \\
           &  (comp. rate) & C4 & Wiki  &  (↑) & (GB, ↓)&  (comp. rate) & C4 & Wiki &  (↑) & (GB, ↓) \\
           \midrule
           \midrule
           Baseline & 16 & 7.08 & 5.68    & 1$\times$ & 12.7 & 16 & 7.08 & 5.68  & 1$\times$ & 12.7 \\
           \midrule
           RTN 
               & 3 (5.33) & 28.26 & 25.61  &2.3$\times$ & 2.9
               & 4 (4.00) & 7.73	& 6.29  & 2.0$\times$& 3.7 \\
           GPTQ
                & 3 (5.33)  & 9.55 & 7.55  & 2.3$\times$ & 2.9
                 & 4 (4.00)  & 7.43	& 5.94   & 2.0$\times$ & 3.7 \\
            SpQR  & - & - & -  & - & -  & 3.94 (4.06) & 7.28 & 5.87  & 1.2$\times^\dagger$ & N/A \\
           \hc \OURS  
                     & 3.02 (5.29) & \textbf{7.75} &\textbf{6.32}  & 2.1$\times$& 2.9 
                     &  4.05 (3.95) & \textbf{7.21} &\textbf{5.79}   & 1.8$\times$& 3.8\\
           \midrule
           GPTQ (g128, no reorder)$^\ddagger$ 
                       & 3.24 (4.93)  & 10.09 & 8.85  & 2.0$\times$ & 3.0 
                       & 4.24 (3.77)  & 7.80 & 6.07  & 1.6$\times$& 3.8 \\
           GPTQ (g128)$^\ddagger$  
                       & 3.24 (4.93)  & 7.89 & 6.27  & 0.2$\times$ & 3.0 
                       & 4.24 (3.77)  & 7.21 & 5.78   & 0.4$\times$& 3.8 \\
           AWQ (g128) 
                      & 3.24 (4.93)  & 7.90 & 6.44  & 2.0$\times$ & 3.0 
                      & 4.24 (3.77)  &7.22 & 5.82  & 1.6$\times$& 3.8 \\
           \hc \OURS (0.45\%)
                             & 3.24 (4.93) & \textbf{7.56} & \textbf{6.13}  & 1.9$\times$ & 3.1 
                             &  4.27 (3.75) & \textbf{7.18} & \textbf{5.77}   & 1.7$\times$ & 4.0 \\

        \bottomrule
        
        \end{tabular}
        }

     \end{subtable}
         \begin{subtable}
        \centering
        \scriptsize{
        \vspace{-1mm}
        \begin{tabular}{c|c|cc|cc|c|cc|cc}
        \toprule
            {\textbf{LLaMA-13B}} & \multicolumn{5}{c|}{\textbf{3-bit}}  & \multicolumn{5}{c}{\textbf{4-bit}} \\
            \midrule
           \multirow{2}{*}{\textbf{Method}} & \textbf{Avg. Bits} & \multicolumn{2}{c|}{\textbf{PPL} (↓)} & {\textbf{Speedup}} & {\textbf{Mem.} }  & \textbf{Avg. Bits} & \multicolumn{2}{c|}{\textbf{PPL} (↓)} & {\textbf{Speedup}} & {\textbf{Mem.} } \\
           &  (comp. rate) & C4 & Wiki  &  (↑) & (GB, ↓)&  (comp. rate) & C4 & Wiki &  (↑) & (GB, ↓) \\
           \midrule
           \midrule
           Baseline & 16 & 6.61 & 5.09   & 1$\times$& 24.6& 16 & 6.61 & 5.09 & 1$\times$ & 24.6\\
           \midrule
           RTN 
               & 3 (5.33) & 13.24 & 11.78 & 2.7$\times$ &5.3
               & 4 (4.00) & 6.99 & 5.53 & 2.3$\times$ & 6.8\\
           GPTQ
                & 3 (5.33)  & 8.22 & 6.22  & 2.7$\times$ & 5.3 
                 & 4 (4.00)  & 6.84 & 5.29  & 2.3$\times$& 6.8\\
            SpQR & - & - & -  & - & - & 3.96 (4.04) & 6.72 & 5.22  & 1.2$\times^\dagger$ & N/A \\
           \hc \OURS
                     & 3.02 (5.30) & \textbf{7.08} &\textbf{5.60}  & 2.4$\times$ & 5.4
                      &  4.04 (3.96) & \textbf{6.71} &\textbf{5.18}  & 2.0$\times$ & 6.9 \\
           \midrule
           GPTQ (g128, no reorder)$^\ddagger$ 
                       & 3.25 (4.92)  & 7.16  &  5.53 & 2.2$\times$ & 5.7 
                        & 4.25 (3.77)  &  6.71  &  5.18  & 1.9$\times$ & 7.2\\
           GPTQ (g128)$^\ddagger$ 
                       & 3.25 (4.92)  & 7.12 & 5.47 & 0.2$\times$ & 5.6 
                        & 4.25 (3.77)  & 6.70 & 5.17  & 0.4$\times$ & 7.0\\
           AWQ (g128)
                      & 3.25 (4.92)  & 7.08 & 5.52 & 2.2$\times$ & 5.7 
                       & 4.25 (3.77)  & 6.70 & 5.21 & 1.9$\times$ & 7.2 \\
           \hc \OURS (0.45\%)
                             & 3.24 (4.94) & \textbf{6.92} & \textbf{5.45}  & 2.2$\times$ & 5.8
                             &  4.26 (3.76) & \textbf{6.68} & \textbf{5.17}  & 1.9$\times$ & 7.3\\

        \bottomrule
        \end{tabular}
        
     \vspace{-1mm}
        }
    
     \end{subtable}

     }
     }
     }
\end{table*}

\begin{figure*}[t!]
\centering{
\centerline{
  \includegraphics[width=\textwidth]{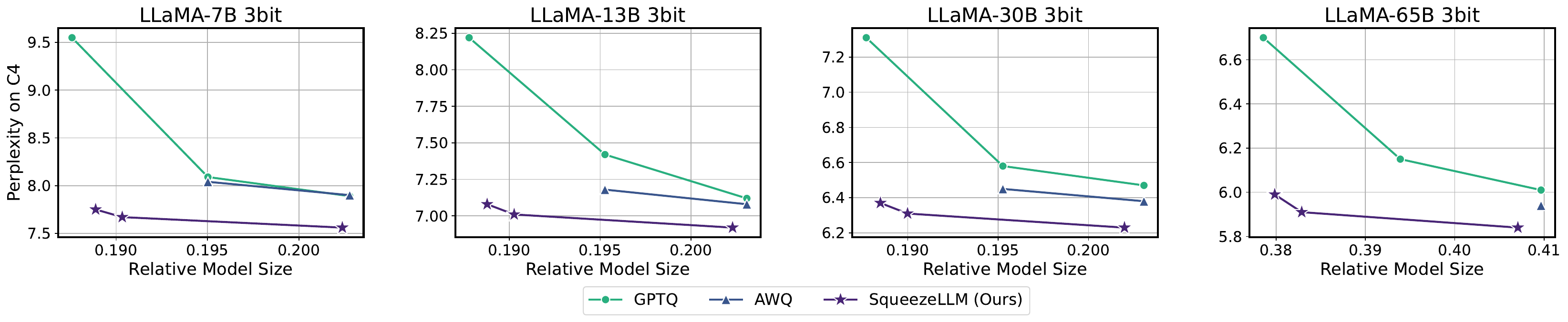}
  }
  \vspace*{-4mm}
  \caption{
Perplexity comparison PTQ methods for 3-bit LLaMA quantization, evaluated on C4.
The x-axes are the relative model sizes with respect to the model size in FP16. 
Different size-perplexity trade-offs are achieved by adjusting the group size for GPTQ and AWQ and the sparsity level for ours.
Our quantization method consistently and significantly outperforms GPTQ and AWQ across all model size regimes, with a more pronounced gap in lower-bit and smaller model sizes.
\label{fig:main_results}
}
}
\vspace{-2mm}
\end{figure*}

Tab.~\ref{tab:llama-7b-main} shows quantization results for LLaMA along with the baseline methods. 
The models are grouped based on their size to better compare size-perplexity trade-offs.
See Fig.~\ref{fig:main_results} for a visual illustration.
Below we use LLaMA-7B as the main example to discuss the impact of dense-only and Dense-and-Sparse quantization, and we then discuss how these trends extend to larger models.
We provide the full evaluation result on all LLaMA models in Tab.~\ref{tab:llama-7b-main-full}.

\noindent
\textbf{Dense-only Quantization.}
In Tab.~\ref{tab:llama-7b-main} (Top), we compare dense-only \OURS with 0\% sparsity level and GPTQ without grouping.
With 4-bit quantization, our method exhibits minimal degradation compared to the FP16 baseline, with only $\sim$0.1 perplexity degradation on C4 and WikiText2, while reducing the model size by 3.95$\times$.
Moreover, when compared to non-grouped GPTQ  
our method shows significant perplexity improvement of up to 0.22.

The performance gap between the two methods becomes more pronounced with 3-bit quantization. 
\OURS  outperforms GPTQ by a substantial margin of 1.80/1.22 points on C4/WikiText2 with a 5.29$\times$ compression rate. 
This is only 0.67/0.55 points off from the FP16 baseline.
This demonstrates the effectiveness of the sensitivity-based non-uniform method for ultra-low-bit quantization.

\blfootnote{{{$^\dagger$Since SpQR does not release their kernel implementation, we conduct our best-effort comparison using their reported speedup numbers. See Appendix~\ref{appendix:experiment_setup} for details.} }}
\blfootnote{{{$^\ddagger$}GPTQ with activation ordering  incurs a significant latency penalty as elements in the same channel are associated with different scaling factors, resulting in distributed memory accesses (Sec.~\ref{subsec:deployment}). GPTQ \textit{without} activation ordering alleviates the latency issue at the cost of a substantial perplexity degradation.}}

\vspace{-0.5mm}
\noindent
\textbf{Dense-and-Sparse Quantization.}
By leveraging the Dense-and-Sparse quantization, we achieve a further reduction in the perplexity gap from the FP16 baseline, as shown in Tab.~\ref{tab:llama-7b-main}.
This improvement is particularly significant with 3-bit quantization, where extracting just 0.45\% of the values yields around 0.2 perplexity improvement.
This enables nearly lossless compression with less than 0.1/0.5 perplexity deviation from the FP16 baseline for 4/3-bit, respectively.

Both GPTQ and AWQ use a grouping strategy to enhance performance with a slight overhead in model size.
However, we demonstrate that \OURS with a sparsity level of 0.45\% consistently outperforms both GPTQ/AWQ with a group size of 128 in all scenarios with comparable model sizes.
This is more pronounced for 3-bit quantization, where \OURS with a 0.45\% sparsity level outperforms both GPTQ/AWQ with a group size of 128 by up $\sim$0.3 perplexity.

\begin{table*}
\caption{Comparison of PTQ methods on zero-shot MMLU accuracy applied to Vicuna v1.1 and v1.3.
We add peak memory usage in GB for comparison.
Additional results on 5-shot MMLU evaluation can be found in Appendix~\ref{appendix:additional-mmlu-5shot}.
}\label{tab:mmlu}
\vspace{1mm}
\centering
\scriptsize{
\setlength{\tabcolsep}{4.5pt}{
\begin{tabular}{c|c|cc|cc|cc|cc|cc}
        \toprule
        \multirow{2}{*}{\textbf{Method}} &  \textbf{Avg.}& \multicolumn{2}{c|}{\textbf{Vicuna-7B (v1.1)}}  & \multicolumn{2}{c|}{\textbf{Vicuna-13B (v1.1)}}  & \multicolumn{2}{c|}{\textbf{Vicuna-7B (v1.3)}}  & \multicolumn{2}{c|}{\textbf{Vicuna-13B (v1.3)}}  & \multicolumn{2}{c}{\textbf{Vicuna-33B (v1.3)}}\\
         & \textbf{bit} & Acc (↑)& Mem (GB, ↓)  & Acc (↑) & Mem (GB, ↓)  & Acc (↑) & Mem (GB, ↓)  & Acc (↑) & Mem (GB, ↓)  & Acc (↑) & Mem (GB, ↓) \\
        \midrule
        \midrule        
        Baseline & 16 & 39.1\% & 12.7 & 41.2\%  & 24.6 &  40.2\% & 12.7 & 43.3\%  & 24.6 & 49.5\% & OOM\\
        \midrule
        AWQ (g128) & 4.25 & 38.0\% & 3.8 & 40.4\%  & 7.2 & \textbf{39.6}\% & 3.8 & 42.2\%  & 7.2 & 49.5\% & 17.2 \\
        \hb \OURS  & 4.05 & 38.8\% & 3.8 & 39.2\%  & 6.9 & 39.3\% & 3.8 & \textbf{44.1}\%  & 6.9 & 48.0\% & 17.5\\
        \hc \OURS (0.45\%) & 4.26 & \textbf{39.4\%} & 4.0 & \textbf{41.0\%}  & 7.3 & 39.5\% & 4.0 & 43.8\%  & 7.3 & \textbf{49.9\%} & 18.7\\
        \midrule
        AWQ (g128) & 3.25 & 36.5\% & 3.0 & 37.6\%  & 5.7 & 37.4\% & 3.0 & 40.7\%  & 5.7  & 46.4\% & 13.2\\
        \hb \OURS  & 3.02 & 36.0\% & 2.9 &37.2\%  & 5.4 & 35.1\% & 2.9 & 40.5\%  & 5.4 & 46.2\% & 12.5\\
        \hc \OURS (0.45\%) & 3.24 & \textbf{37.7\%} & 3.1 & \textbf{39.4\%}  & 5.8  & \textbf{37.6\%} & 3.1 & \textbf{40.8\%}  & 5.8 & \textbf{47.7\%} & 14.7\\
        \bottomrule
\end{tabular}
}
}
\end{table*}

\vspace{-0.5mm}
\noindent
\textbf{Results on Larger Models.} 
In Tab.~\ref{tab:llama-7b-main} (13B) and Tab.~\ref{tab:llama-7b-main-full} (30/65B), we observe that the trend in 7B extends to larger models, where \OURS consistently outperforms other PTQ methods across all models and bit widths. 
Such a trend is also illustrated in Fig.~\ref{fig:main_results} for 3-bit quantization where even \textit{dense-only} \OURS achieves comparable perplexity to \textit{grouped} GPTQ/AWQ. 
With sparsity, we can further improve perplexity, reducing the gap from the FP16 baseline to less than 0.1/0.4 perplexity points for 4/3-bit quantization.
Notably, with 3-bit quantization, our approach achieves up to a 2.1$\times$ reduction in perplexity gap from the FP16 baseline compared to existing methods.
Further ablation studies on our design choices are provided in Appendix~\ref{subsec:ablation}, and additional results on the LLaMA2 and OPT models can be found in Appendix~\ref{appendix:additional-results}.

\vspace{-1mm}
\subsection{Quantization of Instruction Following Models}
\label{subsec:instruction-following}
\vspace{-1mm}

Instruction tuning has emerged as a method for improving the model's ability to respond to user commands. 
We explore the quantization of instruction-following models
to demonstrate the benefits of \OURS  in terms of accuracy preservation by applying it to the Vicuna models, and evaluating the performance with the following approaches.

\noindent
\textbf{MMLU Evaluation.}
We first evaluate the baseline and quantized models on the MMLU benchmark where the weighted accuracy in the zero-shot setting is provided in Tab.~\ref{tab:mmlu} for Vicuna models.
As we can see, 3-bit \OURS achieves higher accuracy for all models compared to AWQ and also preserves the FP16 baseline accuracy with 4-bit quantization. 
5-shot results are provided in Appendix~\ref{appendix:additional-mmlu-5shot}.

\begin{figure*}[t!]
\centering{
\centerline{
  \includegraphics[width=0.85\linewidth]{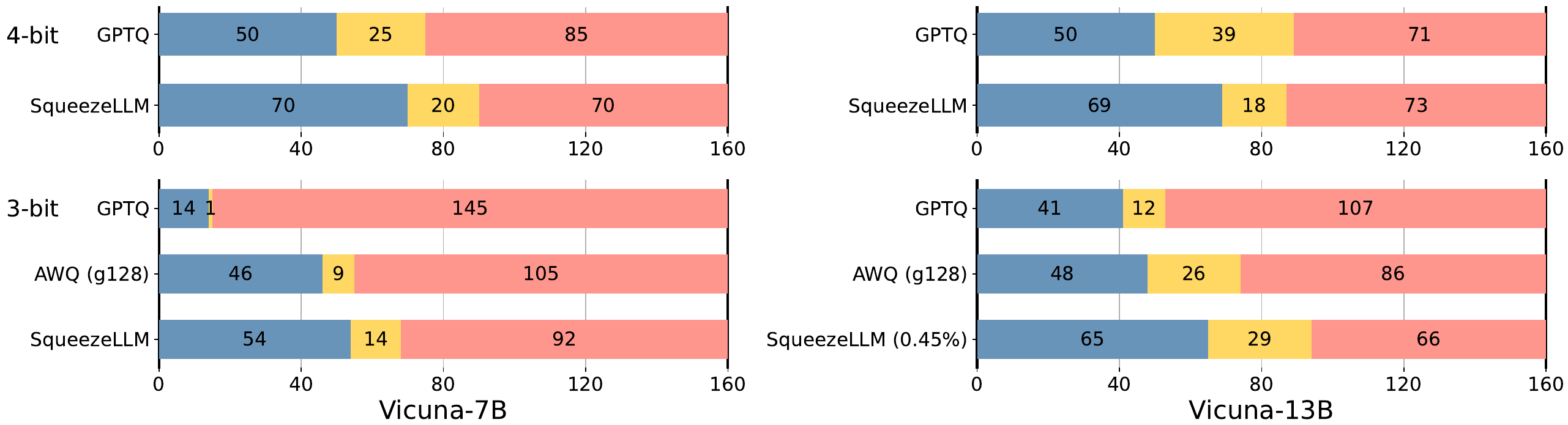}
  }
  \vspace*{-4mm}
  \caption{
 Comparison of PTQ methods applied to Vicuna v1.1. Blue / yellow / red represent the number of times that the quantized model won / tied / lost against the baseline FP16  model.
  This evaluation was performed using the methodology from Vicuna.
 \label{fig:instruction-following}
}
\vspace{-2mm}
}
\end{figure*}

\noindent
\textbf{Instruction-Following Ability.}
Another approach for evaluating instruction-following ability is to ask GPT-4 to rank the generated responses
as presented in~\cite{vicuna2023}.
As shown in Fig.~\ref{fig:instruction-following},
\OURS without sparsity achieves near-perfect performance (i.e., 50/50 split) with 4-bit quantization for both Vicuna-7B and 13B, outperforming GPTQ with the same model size.
In the case of 3-bit quantization, \OURS outperforms both GPTQ and AWQ with comparable model sizes. In the case of the Vicuna-13B model, achieving a near-perfect 50/50 split for 3-bit quantization.

\subsection{Hardware Deployment and Profiling}
\label{subsec:deployment}

\begin{table*}[t!]
\caption{
Latency (s) and peak memory usage (GB) of 3-bit LLaMA when generating 128 tokens on an A6000 GPU.
The table compares the FP16 baseline, non-grouped and grouped GPTQ with activation ordering, and \OURS with different sparsity levels.
For comparison, we include bitwidth and perplexity on the C4 benchmark. See Tab.~\ref{tab:hardware-deployment-seqlen1024} for additional results on generating 1024 tokens, and see Tab.~\ref{tab:hardware-a100} for additional benchmarking results on an A100 GPU.
}
\vspace{-2mm}

\label{tab:hardware-deployment-seqlen128}
\centering{
\footnotesize{
\setlength{\tabcolsep}{4pt}{
    \hspace{-8mm}
    
    \begin{subtable}
        \centering
        \scriptsize{
        \vspace{-1mm}
        \begin{tabular}{c|c|ccc|ccc|ccc|ccc}
        \toprule
           \multirow{2}{*}{\textbf{Method}} & {\textbf{Bit}} & \multicolumn{3}{c|}{\textbf{7B}} &  \multicolumn{3}{c|}{\textbf{13B}} & \multicolumn{3}{c|}{\textbf{30B}} & \multicolumn{3}{c}{\textbf{65B}} \\
           & \textbf{width} & PPL (C4) & Lat (s) & Mem (G)& PPL (C4) & Lat (s) & Mem (G)& PPL (C4) & Lat (s) & Mem (G) & PPL (C4) & Lat (s) & Mem (G)\\
           \midrule
           \midrule
           Baseline & 16 & 7.08 & 3.2 & 12.7 & 6.61 & 5.6 & 24.6 & 5.98  & OOM & OOM & 5.62 & OOM & OOM  \\
           \midrule
           GPTQ & 3 & 9.55 & 1.4 & 2.9 & 8.22 & 2.1 & 5.3 & 7.31 & 4.0 & 12.3 & 6.70 & 6.7 & 24.0 \\
            \hc \OURS & 3.02 & 7.75 & 1.5 & 2.9 & 7.08 & 2.4 & 5.4 & 6.37 & 4.0 & 12.5 & 5.99 & 7.6 & 24.5 \\
            \midrule
           GPTQ (g128) & 3.25 & 7.89 & 13.7 & 3.0 & 7.12 & 24.2 & 5.6 & 6.47 & 61.9 & 12.9 & 6.01 & 117.8 & 25.1\\
           
           \hc \OURS (0.45\%) & 3.24 & 7.56 & 1.7 & 3.1 & 6.92 & 2.5 & 5.8 & 6.23 & 4.4 & 14.7 & 5.84 & 8.8 & 28.0\\
        \bottomrule
\end{tabular}
}
       
    \end{subtable}

     }
     }
     }
\end{table*}

We show the latency and peak GPU memory usage of \OURS in Tab.~\ref{tab:hardware-deployment-seqlen128} 
on an A6000 GPU for different configurations when generating 128 tokens.
We observe that the LUT-based non-uniform approach in \OURS (3rd row) shows up to 2.4$\times$ speedup compared to the FP16 baseline, and exhibits comparable latency and peak memory usage to the uniform quantization of non-grouped GPTQ (2nd row). 
This indicates that the overhead associated with LUT-based dequantization is small,
especially considering the significant perplexity gains it enables.

Additionally, when incorporating sparsity, we still observed latency gains relative to the FP16 baseline.
As shown in Tab.~\ref{tab:hardware-deployment-seqlen128}, keeping 0.45\% of parameters in FP16 (4th row)
only adds around 10\% latency overhead relative to the dense-only implementation, while still resulting in up to $2.2\times$ speed up compared to the FP16 baseline.
In contrast, when accounting for permutation, the GPTQ runtime is degraded heavily (5th row). 
This latency penalty is due to permutation, which means that elements in the same channel need to be scaled using different scaling factors (which are accessed using group indices); it is challenging for these distributed memory accesses to be performed efficiently, as GPUs rely heavily on coalesced memory accesses in order to optimally use memory bandwidth. 
This shows how our Dense-and-Sparse quantization methodology allows for both higher accuracy as well as better performance relative to GPTQ.
Additional evaluation results on generating 1024 tokens are provided in Tab.~\ref{tab:hardware-deployment-seqlen1024}, where we observe a similar trend.

\section{Conclusion}

We have presented \OURS which attempts 
to address the Memory Wall problem associated with generative LLM inference that is memory-bound. 
\OURS incorporates two novel ideas that allow ultra-low precision quantization of LLMs with negligible degradation in generation performance:
the sensitivity-based non-uniform quantization method; 
and the Dense-and-Sparse decomposition that resolves the outlier issue.
We have evaluated \OURS on a wide range of models and datasets that assess language modeling, problem-solving, and instruction-following capabilities of quantized models, where we have demonstrated that our quantization method can consistently outperform the previous state-of-the-art methodologies.

\subsection*{Impact Statement}
This paper introduces advancements in machine learning through a method that results in more lightweight models by improving computational efficiency. While this technique will enable broader access to the applications of machine learning technologies across diverse sectors, we do not foresee direct negative social consequences that require specific highlights. Our work aims at fostering innovation and inclusivity in the field, making advanced technologies more available to a wider range of users and developers.

\subsection*{Acknowledgements}
The authors would like to acknowledge Karttikeya Mangalam, Nicholas Lee, and Thanakul Wattanawong for helpful discussions and brainstorming.
We acknowledge gracious support from 
Google Cloud, Google TRC team, and specifically Jonathan Caton, Jing Li, Jiayu Ye, and Prof. David Patterson.
Prof. Keutzer's lab is sponsored by Intel corporation, Intel VLAB team, Intel One-API center of
excellence, as well as gracious funding from Furiosa, Berkeley Deep Drive, and BAIR.
Michael W. Mahoney would also like to acknowledge
a J. P. Morgan Chase Faculty Research Award 
as well as 
the DOE, NSF, and IARPA.
Sehoon Kim would like
to acknowledge the support from the Korea Foundation for
Advanced Studies (KFAS).
Our conclusions do not necessarily reflect the position or the policy of our sponsors, and no official endorsement should be~inferred.

\bibliography{paper}
\bibliographystyle{icml2024}

\renewcommand\thefigure{\thesection.\arabic{figure}} 
\setcounter{figure}{0} 

\renewcommand\thetable{\thesection.\arabic{table}} 
\setcounter{table}{0} 
\newpage

\clearpage
\appendix
\onecolumn

\section{Related Works on Transformer Quantization  }
\label{appendix:related}
Quantization methods can be broadly categorized based whether retraining is required or not~\cite{gholami2021survey}.
Quantization-Aware Training (QAT) requires retraining the model
to adapt its weights to help recover accuracy after quantization~\cite{zafrir2019q8bert,shen2020q,kim2021bert,zhang2023qd,zhang2020ternarybert, bai2020binarybert}, whereas Post-Training Quantization (PTQ)
does not involve retraining~\cite{zhao2019improving, cai2020zeroq, shomron2021post, oh2022non, li2023q}.
While QAT often results in better accuracy, it is often infeasible for LLMs due to the expensive retraining cost and/or lack of access to the training data and infrastructure. As such, most works on LLM quantization
have focused on PTQ~\cite{yao2022zeroquant,dettmersgpt3, frantar2022gptq, xiao2023smooth, yuan2023rptq, lin2023awq}.
Our work also focuses on the PTQ approach.

Quantization methods can be also classified as uniform or non-uniform~\cite{gholami2021survey}. 
Uniform quantization~\cite{frantar2022gptq,lin2023awq,dettmers2023spqr,zafrir2019q8bert, shen2020q, kim2021bert, huang2023output, liu2023noisyquant}, 
which uniformly divides weight ranges into bins, has gained popularity since it allows faster computation by using quantized precision arithmetic.
 However, recent hardware trends indicate that faster computation does not necessarily translate to improved end-to-end latency or throughput~\cite{aimemory}, particularly in memory-bound tasks like generative LLM inference (Sec.~\ref{sec:memorywall}). 
Furthermore, uniform quantization can be sub-optimal when the weight distribution is non-uniform, as in  LLMs (Fig.~\ref{fig:nonuniform}).

Hence, we focus on non-uniform quantization, which non-uniformly allocates quantization bins without constraints for a more accurate representation of weights and smaller quantization errors.
While it does not support integer arithmetic for computational acceleration, 
this drawback is not significant for memory-bound problems, as in our focus, where the primary bottleneck lies in memory bandwidth rather than computation.
Among non-uniform quantization methods~\cite{jeon2022mr, chung2020extremely}, the most similar work to ours is GOBO~\cite{zadeh2020gobo}, which introduces a k-means clustering-based look-up table approach.
Our work introduces two novel methods as compared to GOBO: (i) sensitivity-based methods; and (ii) Dense-and-Sparse quantization methodologies, which yield substantial improvements within the k-means-based non-uniform quantization framework.

\section{Experiment Setup (Details)}
\label{appendix:experiment_setup}
\textbf{Models and Datasets.} 
We have conducted comprehensive evaluations of \OURS using various models on different tasks.
First, in the language modeling evaluation, we apply \OURS to the LLaMA~\cite{touvron2023llama}, LLaMA2~\cite{touvron2023llama2} and OPT~\cite{zhang2022opt} models and measure the perplexity of the quantized models on the  C4~\cite{raffel2020exploring} and WikiText2~\cite{merity2016pointer} datasets with a chunk size of 2048. 
We also evaluate the domain-specific knowledge and problem-solving ability through the MMLU benchmark~\cite{hendryckstest2021} using the instruction-tuned Vicuna (v1.1 and v1.3) models. 
We used the Language Model Evaluation Harness to run zero-shot evaluation across all tasks~\cite{eval-harness}. 
Finally, we evaluate the instruction following ability following the methodology presented in~\cite{vicuna2023}.
To do so, we generate answers for 80 sample questions and compared them to the answers generated by the FP16 counterpart using the GPT-4 score. To minimize the ordering effect, we provide the answers to GPT-4 in both orders, resulting in a total of 160~queries.

\textbf{Baseline Methods.} We compare \OURS against PTQ methods for LLMs including RTN as well as state-of-the-art methods including GPTQ~\cite{frantar2022gptq}, AWQ~\cite{lin2023awq} and SpQR~\cite{dettmers2023spqr}.
To ensure a fair comparison, we use GPTQ \textit{with} activation ordering throughout all experiments unless specified, which addresses the significant performance drop that would otherwise occur.
For AWQ, we use official quantized models or reproduce using their official code if they are not available except for LLaMA 65B with group size 256, which ran into OOM even on A100-80G.
Evaluations are then conducted based on our perplexity method. 
For SpQR, we rely on the paper's reported numbers since their perplexity evaluation methodology is identical to ours.
SpQR aims to enhance 3-bit and 4-bit models by introducing grouping, bi-level quantization, and sparsity, making them approximately 4 and 4.6 bits on average for LLaMA. 
In contrast, \OURS aims to preserve 3 and 4-bit as closely as possible, minimizing any extra model size overhead.
Therefore, we present our best-effort comparison of SpQR and \OURS by comparing 3-bit SpQR models, which average around 4 bits, and our 4-bit models, both of which possess similar model sizes. 

\textbf{Latency Profiling. }
We measure the latency and peak memory usage for generating 128 and 1024 tokens on an A6000 machine using the Torch CUDA profiler.
As an official implementation of GPTQ (in particular, the grouped version) 
is not available, we implement an optimized kernel for single-batch inference based on the most active open-source codebase~\cite{gptq-for-llama}.

To compare latency with SpQR, we rely on their reported speedup numbers to make our best-effort comparison, as their kernel implementation is not publicly available. 
Regarding AWQ, we use the GPTQ kernel without activation ordering since they exhibit identical behavior during inference. Although AWQ has released their own kernel implementation, their 3-bit kernels are not publicly available. Furthermore, they have incorporated optimizations that are unrelated to quantization, such as LayerNorm and positional embedding, which are universally applicable to other quantization methods. To ensure a fair comparison with other methods, we refrained from using their released kernels.

\section{Data Skew in Per-channel Sparsity Pattern}
\label{subsec:data_skew}

\begin{figure*}[t!]
\centering{
\centerline{
  \includegraphics[width=0.65\textwidth]{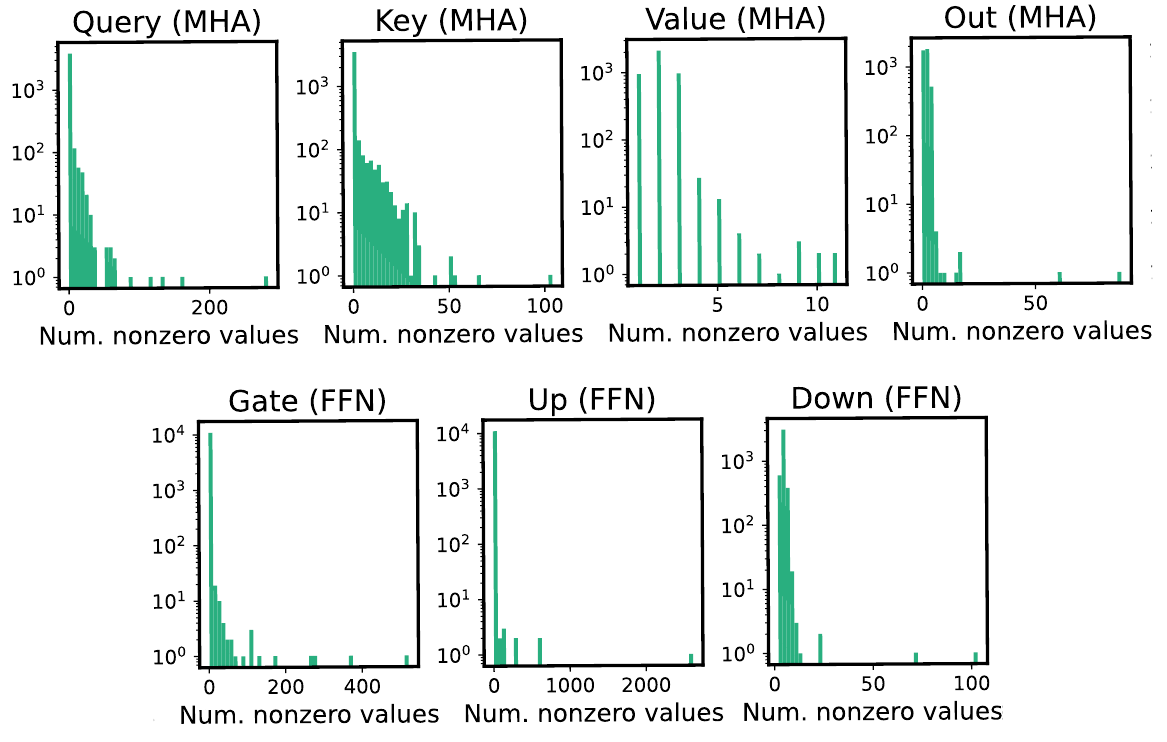}
  }
  \caption{
    Histograms of the number of non-zero entries per output channel in 7 different linear layers in the first LLaMA-7B block. 
    The histograms reveal the presence of a few channels that contain significantly more non-zero entries than others, highlighting the skew in the sparsity patterns across different channels within the linear layers.
  }
\label{fig:data-skew}
}
\end{figure*}

\begin{table}[t!]
\caption{
Hardware profiling of latency and memory usage using different kernel implementations for LLaMA 7B, 13B, 30B, and 65B quantized into 3-bit when generating 128 tokens on an A6000 GPU.
The first row shows the performance of \OURS without sparsity as a reference. 
The second row shows the performance of \OURS with a sparsity level of 0.45\% using a standard kernel for processing a CSR matrix. 
The third row shows the performance of \OURS with a sparsity level of 0.45\% using a balanced sparse kernel that allocates 10 nonzeros per thread, thereby more efficiently handling skewed sparse matrices.
}
\label{tab:hardware-topK}
\centering{
\footnotesize{
\setlength{\tabcolsep}{9pt}{

    \hspace{-8mm}
    
    \footnotesize
        \centering
        \vspace{-1mm}
        \begin{tabular}{c|c|cccc|cccc}
        \toprule
           \textbf{Sparse} & \textbf{Sparsity}& \multicolumn{4}{c|}{\textbf{Latency (Seconds)}} &  \multicolumn{4}{c}{\textbf{Peak Memory (GB)}} \\
           \textbf{Kernel}&\textbf{Level}& 7B & 13B & 30B & 65B &  7B & 13B & 30B  & 65B \\
           \midrule
           \midrule
           - & 0\%& 1.5  & 2.4  & 4.0 & 7.6 & 2.9 & 5.4 & 12.5 & 24.5\\
           \midrule
           {\textbf{Standard}} & 0.45\% & 3.9 & 6.2 & 12.5 & 14.4 & 3.2 & 5.8 & 13.7 & 28.0 \\
          {\textbf{Balanced}} & 0.45\% & 1.7  & 2.6  & 4.4 & 8.8 & 3.1 & 5.8 & 14.7 & 28.0 \\
        \bottomrule
\end{tabular}
    \vspace{3mm}
     }
     }
     }
\end{table}


Fig.~\ref{fig:data-skew} provides the distribution of nonzero entries per output channel across different linear layers in the first LLaMA-7B block. 
This plot shows that the nonzero distribution is heavily skewed, with a few channels containing a much larger proportion of nonzero values.
This skewed distribution makes it challenging to efficiently perform computations using the sparse matrix, as it is difficult to distribute the nonzero elements evenly across parallel processing units.
This motivates our modified kernel for handling channels with a large number of outliers in order to reduce the runtime overhead of the sparse matrices. 
As outlined in Tab.~\ref{tab:hardware-topK}, we observed over 100\% added runtime overhead when employing a standard CSR-based kernel. 
However, if we allocate each thread to process a fixed number of nonzeros (rather than having each thread process an entire row) we were able to drastically reduce the runtime overhead to 10-20\% with both sensitive values and outliers.

\section{Ablation Studies}
\label{subsec:ablation}

\subsection{{Sensitivity-Based Quantization.}}

\begin{table}[t!]
\caption{
Ablation study comparing sensitivity-agnostic and sensitivity-based non-uniform quantization on the LLaMA-7B model with 3-bit quantization, measured by perplexity on the C4 benchmark. The baseline
model in FP16 achieves a perplexity of 7.08.
}
\label{tab:ablation-sens}
\centering{
\small{
\setlength{\tabcolsep}{7pt}{
    \hspace{-8mm}
    
        \footnotesize
        \centering
        \vspace{-1mm}
        \begin{tabular}{c|cc}
        \toprule
        \textbf{Method} & \textbf{Sensitivity-Agnostic} (↓) & \textbf{Sensitivity-Based} (↓) \\
        \midrule
        \OURS  & 18.08 & \textbf{7.75}\\
           
           \OURS (0.05\%) & 8.10 & \textbf{7.67} \\
           \OURS (0.45\%) & 7.61 & \textbf{7.56} \\
        \bottomrule
\end{tabular}

     }
     }
     }
\end{table}

In our ablation study, we investigate the impact of sensitivity-based weighted clustering on the performance of non-uniform quantization. 
In  Tab.~\ref{tab:ablation-sens}, we compared the performance of sensitivity-based and sensitivity-agnostic approaches in the context of 3-bit quantization of the LLaMA-7B model. 
For sensitivity-agnostic quantization, we apply non-weighted k-means clustering at sparsity levels of 0\%, 0.05\%, and 0.45\%.
The results demonstrate that while non-uniform quantization alone can reduce the perplexity from 28.26 (of RTN uniform quantization) to 18.08 without considering sensitivity, 
incorporating sensitivity-based clustering is critical in reducing the perplexity to 7.75. 
This improvement is consistent across all sparsity~levels.

\subsection{{Impact of Sparsity Levels on \OURS}}
\label{subsec:sparsity}

\begin{figure*}[t!]
\centering{
\centerline{
  \includegraphics[width=0.8\textwidth]{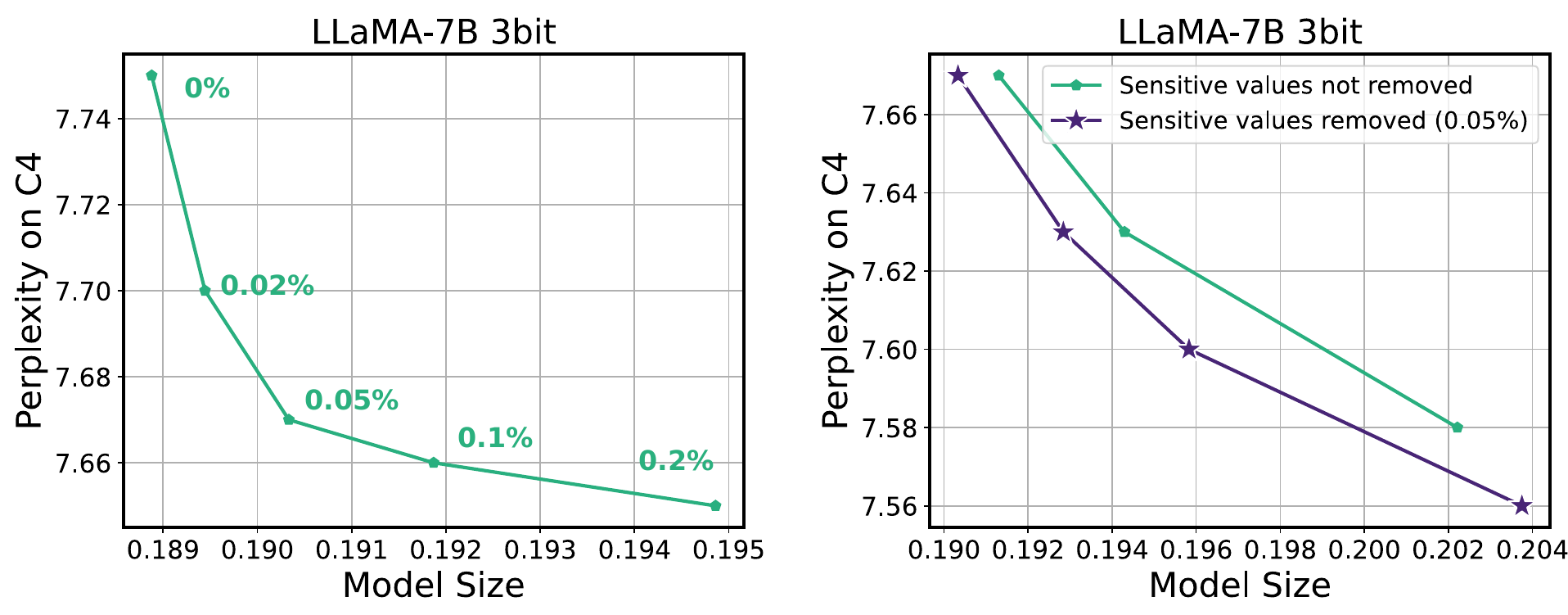}
  }
    \vspace*{-5mm}
\caption{
(Left) Model size (normalized by the size of the FP16 model) and perplexity trade-off with different percentages of sensitive values included in the sparse matrix. Here, no outlier values are included in the sparse matrix.
(Right) Comparison of the performance when the sensitive values are not removed as the sparse matrix (only outlier values are removed) to the case where 0.05\% of the sensitive values are removed. 
In both cases, the trade-offs are obtained by controlling the percentage of outlier values included in the sparse matrix.
  }
\label{fig:ablation-sparsity}
}
\end{figure*}

In Fig.~\ref{fig:ablation-sparsity} (Left), we present the perplexity results of the 3-bit quantized LLaMA-7B model on the C4 benchmarks, with varying percentages of sensitive values extracted as the sparse matrix, ranging from 0\% to 0.2\%. 
The plot demonstrates that the perplexity gain diminishes as the sparsity level of the sensitive values exceeds 0.05\%. 
Therefore, we maintain a fixed sparsity level of 0.05\% for the sensitive values throughout all experiments.

Furthermore, in Figure~\ref{fig:ablation-sparsity} (Right), we compare the performance when the sensitive values are not removed as the sparse matrix (only outlier values are removed) to the case where 0.05\% of the sensitive values are removed. 
In both scenarios, we control the sparsity level by increasing the percentage of outlier values included in the sparse matrix to obtain the trade-off curves. 
The results indicate that the sparsity configuration with both sensitive values and outlier values consistently outperforms the configuration with only outlier values.

\subsection{{Impact of Grouping on \OURS}}
\label{subsec:ablation-grouping}

\begin{figure}[t!]
\centering{
\centerline{
  \includegraphics[width=0.4\textwidth]{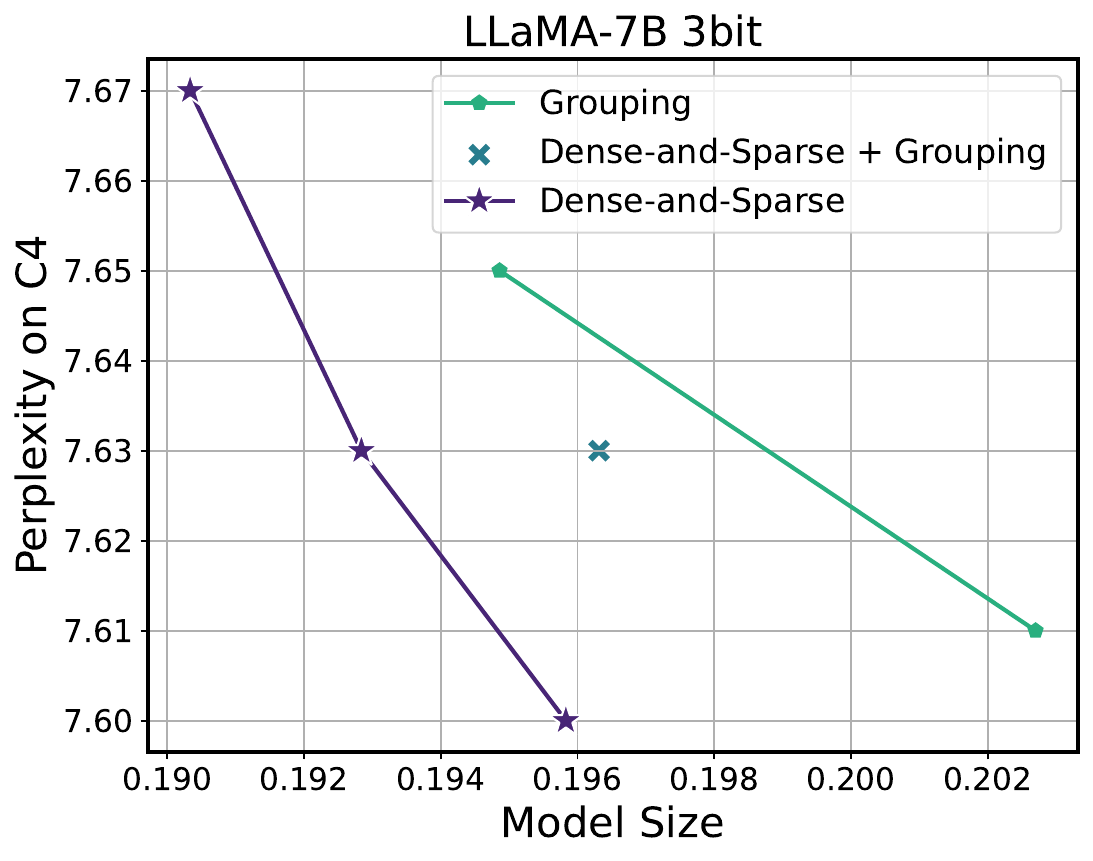}
  }
  \vspace*{-5mm}
  \caption{
Model size (normalized by the size of the FP16 model) and perplexity trade-offs of grouping and the Dense-and-Sparse decomposition on 3-bit quantization of LLaMA-7B.
Here, we compare \OURS with (i) grouping using group sizes of 1024 and 512 (green), (ii) a hybrid approach that combines a group size of 1024 with a sparsity level of 0.05\% (blue), and (iii) the Dense-and-Sparse decomposition approach with varying sparsity levels (violet).
The pure Dense-and-Sparse decomposition always outperforms both grouping and the hybrid approach. 
  }
\label{fig:ablation-grouping}
}
\end{figure}

In Fig.~\ref{fig:ablation-grouping}, we explore the effectiveness of incorporating grouping into \OURS as an alternative approach to improve quantization performance. We compare three configurations: \OURS with (i) grouping using group sizes of 1024 and 512 (green), (ii) a hybrid approach that combines a group size of 1024 with a sparsity level of 0.05\% (blue), and (iii) the Dense-and-Sparse decomposition approach with varying sparsity levels (violet), where 0.05\% of sensitive values are kept and the percentage of outlier values is adjusted.
The results clearly demonstrate that both grouping and the hybrid approach result in suboptimal trade-offs compared to the pure Dense-and-Sparse decomposition approach. 

This can be attributed to two factors. 
First, the Dense-and-Sparse decomposition is a direct solution to the outlier issue. 
In contrast, while grouping can mitigate the impact of outliers to some extent by isolating them within individual groups, it does not provide a direct solution to this issue. 
Second, grouping can introduce significant overhead in terms of storage requirements when combined with non-uniform quantization, since it needs to store one LUT per group.
This can be a considerable overhead compared to the uniform quantization approach, where only a scaling and zero point value per group need to be stored.

\vspace{3mm}
\subsection{{Comparison of Optimization Objectives for Non-uniform Quantization: Minimizing Layer-wise Perturbation versus Final Output Perturbation}}
\label{subsec:ablation-obd}

\begin{figure}[t!]
\centering{
\centerline{
  \includegraphics[width=0.4\textwidth]{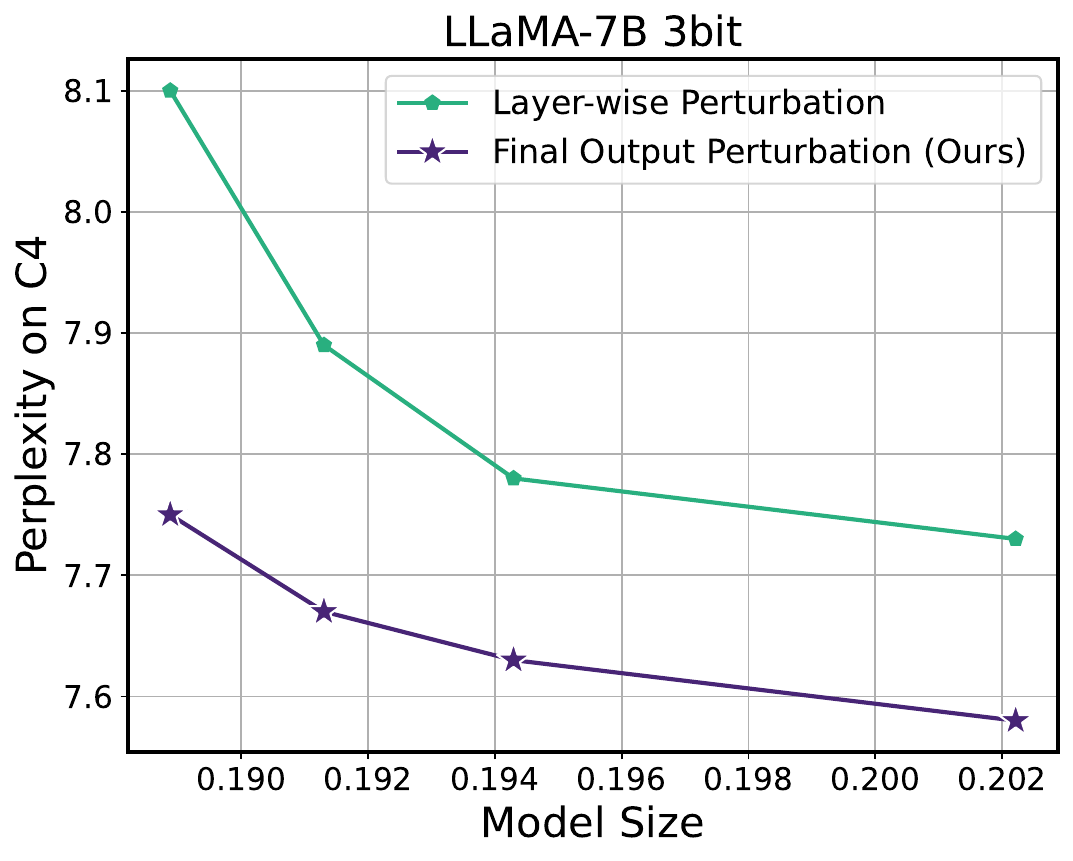}
  }
  \caption{
  Model size (normalized by the size of the FP16 model) and perplexity trade-offs for 3-bit quantization of the LLaMA-7B model using layer-wise perturbation minimization versus final output perturbation minimization as a non-uniform quantization objective.
  The trade-off is obtained by adjusting the sparsity level of the outliers being extracted.
Across all sparsity levels, the OBD framework, which is the foundation for \OURS, consistently outperforms the OBS framework as an alternative approach.
  }
\label{fig:ablation-obd}
}
\end{figure}

While our method targets minimizing the perturbation of the final output of the model during quantization, it is worth noting that minimizing the layer-wise perturbation can also be considered as an alternative.
Most existing solutions for LLM quantization including GPTQ~\cite{frantar2022gptq}, AWQ~\cite{lin2023awq}, and SpQR~\cite{dettmers2023spqr}
have used the latter objective, which aims to minimize the perturbation of output activations in individual layers. 
In this ablation study, we demonstrate that minimizing the final output perturbation is a superior objective to minimizing the layer-wise perturbation. 

When minimizing the layer-wise perturbation, the optimization objective for determining the non-uniform quantization configuration can be reformulated as $\argmin_{Q} \lVert WX - W_QX \rVert_2^2$, where $X$ denotes a batch of input activations. 
This object can be approximated as a weighted k-means clustering problem, where each weight is weighted by the square of the corresponding input activation size.
This indeed results in the activation-based sensitivity/importance metric as in the AWQ framework~\cite{lin2023awq}.

In Fig.~\ref{fig:ablation-obd}, we compare the perplexity on the C4 dataset for 3-bit quantization of the LLaMA-7B model using both objectives.
Across all sparsity levels obtained by adjusting the number of outliers being extracted, \OURS based on final loss perturbation minimization outperforms the alternative of using layer-wise perturbation minimization by a large margin of up to around 0.3 perplexity points.

\begin{table}[t!]
\caption{
Perplexity scores on Wikitext2 for the LLaMA-2 7B model, quantized using non-uniform (\OURS's sensitivity-based quantization) and uniform (RTN) approaches with 3 and 4-bit precision with varying levels of sparsity. 
}
\vspace{1mm}
\label{tab:additional_ablation1}
\centering{
\setlength{\tabcolsep}{7pt}{
    \hspace{-8mm}
        \footnotesize
        \centering
        \begin{tabular}{c|cc|cc}
        \toprule
        \textbf{Bit Width} & \textbf{Sparsity Level (\%)} & \textbf{Avg. Bit Width} & \textbf{Uniform (PPL)} & \textbf{Nonuniform (PPL)}\\
        \midrule
        \midrule
        16-bit &	0	& 16	& 5.47	& \textbf{5.47} \\
        \midrule
        	&0	&4.04	&6.12	&\textbf{5.62} \\
            &0.05	&4.09	&5.95	&\textbf{5.59}\\
        4-bit    &0.45	&4.26	&5.95	&\textbf{5.57}\\
            &2	&5.01	&5.95	&\textbf{5.55}\\
            &4.5	&6.20	&5.94	&\textbf{5.53}\\
            \midrule
           &0	&3.02	&542.00	&\textbf{6.18}\\
            &0.05	&3.07	&27.38	&\textbf{6.05}\\
           3-bit &0.45	&3.24	&26.58	&\textbf{5.96}\\
            &1.5	&3.98	&25.97	&\textbf{5.81}\\
            &4.5	&5.18	&23.58	&\textbf{5.73}\\
        \bottomrule
\end{tabular}

     }
     }
\end{table}

\subsection{Impact of Non-uniform Quantization versus Dense-and-Sparse Decomposition }
\label{subsection:additional_ablation1}
In Tab.~\ref{tab:additional_ablation1}, we perform a detailed analysis to further disambiguate the impact of non-uniform quantization and the Dense-and-Sparse decomposition. 

\vspace{-1mm}
\noindent
\textbf{Uniform vs. Non-uniform Quantization.} 
As can be seen in Tab.~\ref{tab:additional_ablation1}, across all bitwidths and sparsity levels, our non-uniform quantization has noticeable improvements over uniform quantization.

\vspace{-1mm}
\noindent
\textbf{Sparsity Levels.} 
Furthermore, we also report the results with varying sparsity levels of the Dense-and-Sparse decomposition in Tab.~\ref{tab:additional_ablation1}. 
As expected, higher levels of sparsity consistently result in improved performance in any scenario. 
However, there are diminishing returns for larger values of sparse decomposition since only a small portion of the weight values are outliers or sensitive. 
As a consequence, saving additional values into the sparse format does not help as much beyond a certain level, and instead results in higher average bitwidth. 
This is in line with the conclusions in the main experiments where we found a sparsity level of 0.45\% sufficient for the performance gain.

\begin{table}[t!]
\caption{
Perplexity scores on C4 and WikiText2 for the LLaMA-2 7B model, quantized using \OURS with 4-bit and 3-bit with different sparsity level. 
In particular, the sparsity levels of 3-bit quantization are selected to match their average bit widths to that of 4-bit quantization without sparsity.
}
\vspace{1mm}
\label{tab:additional_ablation2}
\centering{
\setlength{\tabcolsep}{7pt}{
    \hspace{-8mm}
        \footnotesize
        \centering
        \begin{tabular}{c|cc|cc}
        \toprule
        \textbf{Bit Width} & \textbf{Sparsity Level (\%)} & \textbf{Avg. Bit Width} & \textbf{C4 (PPL)} & \textbf{WikiText2 (PPL)}\\
        \midrule
        \midrule
        16-bit &	0	& 16	& 6.97	& 5.47 \\
        \midrule
        4-bit	&0	&4.04	& 7.12	&5.62 \\
            \midrule
           \multirow{2}{*}{3-bit} & 1.5	&3.98	&7.35 &5.81	\\
            &2.5	&4.22	&7.32	&5.80\\
        \bottomrule
\end{tabular}
       \vspace{-2mm}

     }
     }
\end{table}

\subsection{Impact of Dense-and-Sparse Decomposition versus Precision}

In Tab.~\ref{tab:additional_ablation2}, we additionally demonstrate that increasing the bit width of the dense component results in higher improvement in perplexity compared to increasing the sparsity level.
Note that 4-bit LLaMA-2 7B model without any sparsity outperforms the 3-bit counterparts with sparsity levels of 1.5\% and 2.5\% that have similar or even larger model sizes.
This observation aligns with the sensitivity level ablation study in Appendix~\ref{subsection:additional_ablation1}, since the Dense-and-Sparse decomposition is only effective to the extent of removing the outliers and sensitive values from the parameters.
Increasing the sparsity level beyond that will not be effective and results in diminishing returns.

\section{Quantization Cost Analysis}

\begin{table}[t!]
\caption{
Peak memory requirement in GB when quantizing different LLaMA models.
}
\vspace{1mm}
\label{tab:e2ememory}
\centering{
\setlength{\tabcolsep}{7pt}{
    \hspace{-8mm}
        \footnotesize
        \centering
        \begin{tabular}{c|c}
        \toprule
        \textbf{Model} & \textbf{Peak Memory (GB)} \\
        \midrule
        LLaMA-7B  & 33\\
           
        LLaMA-13B & 61\\
        LLaMA-30B & 149 \\
        LLaMA-65B & 292 \\
        \bottomrule
\end{tabular}
            \vspace{-2mm}

     }
     }
\end{table}

\subsection{Memory Requirement}
In Tab.~\ref{tab:e2ememory}, we report the memory requirement of \OURS when quantizing different model sizes from 7B to 65B.
Note that our method can have a higher memory requirement than GPTQ.
This is because \OURS performs quantization based on minimizing the perturbation to the loss function of the model which requires computing the Fisher information matrix. 
GPTQ, on the other hand, performs quantization by minimizing the perturbation to the output activation of the individual layer, which does not require back-propagating the gradient through the model to compute the Fisher information matrix. 
However, this is a one-time cost, and as demonstrated below, this gradient computation process is fast, taking only 2-3 minutes even for the largest 65B model.

\begin{table}[t!]
\caption{
End-to-end latency breakdown of quantizing different LLaMA models. Latency is broken down into (i) Fisher information computation on a A100 system and (ii) sensitivity-based k-means clustering on Intel Xeon Gold 6126 with 48 cores.
In the last column, we provide the end-to-end time for GPTQ as reported in the original paper.
}
\vspace{1mm}
\label{tab:e2elatency}
\centering{
\setlength{\tabcolsep}{7pt}{
    \hspace{-8mm}
        \footnotesize
        \centering
        \begin{tabular}{c|cc|c}
        \toprule
        \textbf{Model} & \textbf{Fisher Computation (min)} & \textbf{K-means (min)} & \textbf{GPTQ (min)}\\
        \midrule
        LLaMA-7B  & 0.3 & 11 & 10\\
           
        LLaMA-13B & 0.6 & 17 & 21\\
        LLaMA-30B & 1.3 & 45 & 45\\
        LLaMA-65B & 2.5 & 80 & 96\\
        \bottomrule
\end{tabular}

     }
     }
\end{table}


\subsection{Quantization Time}
In Tab.~\ref{tab:e2elatency}, we additionally assess the end-to-end time for (i) computing the Fisher information on an A100 system and (ii) performing sensitivity-based K-means clustering on Intel Xeon Gold 6126 with 48 cores, which are two major procedures in \OURS. 
Note that the time for computing the Fisher information matrix is minimal, taking only 2.5 minutes with the largest 65B model. 
K-mean clustering can take 11 min for the 7B model and up to 80 min for the 65B model. 
Overall, the computational time requirement of \OURS is on par with that of GPTQ.

\begin{table}[t!]
\caption{
Perplexity on C4 and Wikitext2 of the LLaMA2 7B model after 4-bit quantization, with varying sizes of the calibration dataset used for computing the Fisher information matrix.
}
\vspace{1mm}
\label{tab:dataeff}
\centering{
\setlength{\tabcolsep}{7pt}{
    \hspace{-8mm}
        \footnotesize
        \centering
        \begin{tabular}{c|cc}
        \toprule
        
        \textbf{\# Data Examples} & \textbf{C4} & \textbf{Wikitext2}\\
        \midrule
        1 & 7.89 & 6.41\\
           
        2 & 7.81 & 6.22\\
        5 & 7.73 & 6.20\\
        10 & 7.72 & 6.17\\
        20 & 7.72 & 6.16\\
        100 & 7.72 & 6.18\\
        \bottomrule
\end{tabular}

     }
     }
\end{table}

\subsection{Data Efficiency}

In Tab.~\ref{tab:dataeff}, we provide data efficiency analysis in terms of the number of data samples to calculate the Fisher information matrix (gradients) for sensitivity-based non-uniform quantization.
While we used a calibration set of 100 data samples throughout the paper, a calibration set with as few as 10 examples is typically sufficient to achieve the desired quantization performance. 
Note that both GPTQ and AWQ require 100-200 data points for calibration as reported in the AWQ paper~\cite{lin2023awq}.

\section{Comparison with Other Weight-only Quantization Methods}
In this section, we compare \OURS with more recent weight-only quantization methods including QuIP~\cite{chee2024quip} and OmniQuant~\cite{shao2023omniquant}.

\begin{table}[t!]
\caption{
Perplexity on Wikitext2 of the LLaMA2 13B and 70B models quantized into 4, 3, and 2 bits using \OURS and QuIP~\cite{chee2024quip}.
For QuIP, we use the perplexity numbers that are reported in the original paper as well as our own reproduction using the official codebase. Following the perplexity evaluation method of the QuIP paper, we use sequence length of 4096 (different from other experiments that use sequence length of 2048).
}
\vspace{1mm}
\label{tab:quip}
\centering{
\setlength{\tabcolsep}{7pt}{
    \hspace{-8mm}
        \scriptsize
        \centering
       \begin{tabular}{c|cc|cc}
    \toprule
    \textbf{Model} & \textbf{Config.} & \textbf{Avg. Bit Width} & \textbf{LLaMA2-13B} & \textbf{LLaMA2-70B} \\
    \midrule
    \midrule
    QuIP (original paper) & 4-bit & 4 & - & 3.53 \\
    QuIP (our repr) & 4-bit & 4 & 4.81 & 3.65 \\
    \hb SqueezeLLM & 4-bit & 4.05 & \textbf{4.67} & \textbf{3.21} \\
       \midrule
     QuIP (original paper) & 3-bit & 3 & - & 3.85 \\
    QuIP (our repr) & 3-bit & 3 & 5.25 & 3.84 \\
    \hb SqueezeLLM & 3-bit & 3.02 & \textbf{5.01} & \textbf{3.55} \\
        \midrule
    QuIP (original paper) & 2-bit & 2 & - & 6.33 \\
    QuIP (our repr) & 2-bit & 2 & \textbf{20.54} & \textbf{6.20} \\
    \hb \OURS & 2-bit & 2.01 & 61.25 & 10.86 \\
    \hc \OURS & 2-bit + 0.1\% & 2.05 & \textbf{7.91} & \textbf{5.04} \\
    \hd \OURS & 2-bit + 0.45\% & 2.22 & \textbf{7.43} & \textbf{4.71} \\
    \bottomrule
    \end{tabular}%

     }
     }
\end{table}

\subsection{Comparison with QuIP}

Here, we provide a quantitative comparison of our method to QUIP. 
Given that the QuIP paper only reports performance evaluation of LLaMA2-70B among all LLaMA models, we enrich our comparison by additionally incorporating our own reproduction based on their official codebase.
 Different from other experiments that use sequence length of 2048,
 we use sequence length of 4096, following the perplexity evaluation method of the QuIP paper.
In Tab.~\ref{tab:quip}, we compare the perplexity scores on Wikitext2 for LLaMA2 13B and 70B models quantized to 4, 3, and 2-bit.
Note that we did not include a comparison on LLaMA2 7B as we were unable to achieve reasonable performance with QuIP, as was also reported in~\cite{egiazarian2024extreme}.

 The table indicates that dense-only SqueezeLLM consistently achieves superior performance over QUIP, across all model sizes and quantization bitwidth.
With 2bit quantization, we noticed that solely relying on dense-only quantization may not yield results as competitive as those of QuIP. However, by incorporating just 0.1\% sparsity (additional 0.05 bit; 0.05\% outlier values + 0.05\% sensitive values), SqueezeLLM significantly outperforms QuIP by a considerable margin.

\begin{table}[t!]
\caption{
Perplexity on Wikitext2 of all LLaMA and LLaMA2 models quantized into 4 and 3 bits using \OURS and OmniQuant~\cite{chee2024quip}.
For OmniQuant, we directly use the perplexity numbers that are reported in the original paper.
}
\vspace{1mm}
\label{tab:omni1}
\centering{
\setlength{\tabcolsep}{7pt}{
    \hspace{-8mm}
        \scriptsize
        \centering
       \begin{tabular}{c|cc|ccccccc}
    \toprule
    \textbf{Model} & \textbf{Config.} & \textbf{Avg. Bit Width} & \textbf{7B} & \textbf{13B} & \textbf{30B} & \textbf{65B} & \textbf{2-7B} & \textbf{2-13B} & \textbf{2-70B} \\
    \midrule
\midrule
        Baseline & 16-bit & 16 & 5.68 & 5.09 & 4.1 & 3.53 & 5.47 & 4.88 & 3.32 \\
    \midrule
    Omniquant & 4-bit & 4 & 5.86 & 5.21 & 4.25 & \textbf{3.71} & 5.74 & 5.02 & 3.47 \\
    \hc \OURS & 4-bit & 4.05 & \textbf{5.79 }& \textbf{5.18} &\textbf{4.22} & 3.76 & \textbf{5.62} & \textbf{4.99} & \textbf{3.41} \\
        \midrule
    Omniquant & 4-bit (g128) & 4.24 & 5.77 & 5.17 & 4.19 & \textbf{3.62} & 5.58 & \textbf{4.95} & 3.4 \\
    \hc \OURS & 4-bit (0.45\%) & 4.27 & \textbf{5.77} & \textbf{5.17} & \textbf{4.18} & 3.63 & \textbf{5.57} & 4.96 & \textbf{3.39} \\
    \midrule
    Omniquant & 3-bit & 3 & 6.49 & 5.68 & 4.74 & 4.04 & 6.58 & 5.58 & 3.92 \\
    \hc \OURS & 3-bit & 3.02 & \textbf{6.32} & \textbf{5.60} & \textbf{4.66} & \textbf{4.05} & \textbf{6.18} & \textbf{5.36} & \textbf{3.77} \\
    \midrule
    Omniquant & 3-bit (g128) & 3.24 & 6.15 & \textbf{5.44} & 4.56 & 3.94 & 6.03 & 5.28 & 3.78 \\
    \hc \OURS & 3-bit (0.45\%) & 3.24 & \textbf{6.13} & 5.45 & \textbf{4.44} & \textbf{3.88} & \textbf{5.96} & \textbf{5.23} & \textbf{3.63} \\
    \bottomrule
    \end{tabular}%

     }
     }
\end{table}

\begin{table}[t!]
\caption{
Perplexity on Wikitext2 of all LLaMA2 models quantized into 2 bits using \OURS and OmniQuant~\cite{chee2024quip}.
For OmniQuant, we directly use the perplexity numbers that are reported in the original paper.
}
\vspace{1mm}
\label{tab:omni2}
\centering{
\setlength{\tabcolsep}{7pt}{
    \hspace{-8mm}
        \scriptsize
        \centering
       \begin{tabular}{c|cc|cccc}
    \toprule
    \textbf{Model} & \textbf{Config.} & \textbf{Avg. Bit Width} & \textbf{2-7B} & \textbf{2-13B} & \textbf{2-70B} \\
    \midrule
    \midrule
    Baseline & 16-bit & 16 & 5.47 & 4.88 & 3.32 \\
    \midrule
    OmniQuant & 2-bit & 2 & 37.37 & \textbf{17.21} & \textbf{7.81} \\
    \hb \OURS & 2-bit & 2.01 & \textbf{35.49} & 41.02 & 9.44 \\
    \hc \OURS & 2-bit (0.1\%) & 2.05 & \textbf{13.64} & \textbf{8.56} & \textbf{5.38} \\
    \midrule
    OmniQuant & 2-bit (g128) & 2.24 & 11.06 & 8.26 & 6.55 \\
    \hc \OURS & 2-bit (0.45\%) & 2.22 & \textbf{10.79} & \textbf{7.91} & \textbf{4.99} \\
    \bottomrule
    \end{tabular}%

     }
     }
\end{table}

\subsection{Comparison with OmniQuant}
In Tab.~\ref{tab:omni1}, we compare the perplexity of our method to OmniQuant on WikiText2 using sequence length of 2048.
In particular, the table reports the perplexity numbers of 4 and 3-bit quantized models across all LLaMA and LLaMA2 models. 
For OmniQuant, we directly use the numbers reported in the original paper.
Omniquant and \OURS are grouped in the table so that their model sizes are roughly the same. This comparison demonstrates that \OURS generally outperforms OmniQuant with the same model size and memory constraints.

Additionally, Tab.~\ref{tab:omni2} demonstrates the same comparison using 2-bit quantization.
With 2-bit quantization, the table shows that OmniQuant without grouping outperforms dense-only \OURS on the 13B and 70B models.
This can be attributed to OmniQuant's learnable clipping ranges via a few iterations of training that effectively account for outliers.
\OURS's sensitivity-based nonuniform quantization alone does not inherently address this. 
Handling outliers can be particularly critical for 2-bit quantization where weights should be represented with only four values. 
Nevertheless, introducing a 0.1\% sparsity remarkably enhances SqueezeLLM's performance with a minimal memory overhead increase of 0.05 bit. 
This perplexity improvement is also persistent when comparing OmniQuant with a group size 128 and SqueezeLLM at a 0.45\% sparsity level with roughly the same size.

\section{Additional Hardware Profiling Results}
\label{subsec:additional_hardware_profiling}

\begin{table*}[t!]
\caption{
Latency (s) and peak memory usage (GB) of 3-bit LLaMA when generating 1024 tokens on an A6000 GPU.
The table compares the FP16 baseline, non-grouped and grouped GPTQ with activation ordering, and \OURS with different sparsity levels.
For comparison, we include bitwidth and perplexity on the C4 benchmark. 
}

\label{tab:hardware-deployment-seqlen1024}
\centering{
\footnotesize{
\setlength{\tabcolsep}{3.pt}{
    \hspace{-8mm}
    
    \begin{subtable}
        \centering
        \scriptsize{
        \vspace{-1mm}
        \begin{tabular}{c|c|ccc|ccc|ccc|ccc}
        \toprule
           \multirow{2}{*}{\textbf{Method}} & {\textbf{Bit}} & \multicolumn{3}{c|}{\textbf{7B}} &  \multicolumn{3}{c|}{\textbf{13B}} & \multicolumn{3}{c|}{\textbf{30B}} & \multicolumn{3}{c}{\textbf{65B}} \\
           & \textbf{width} & PPL (C4) & Lat (s) & Mem (G)& PPL (C4) & Lat (s) & Mem (G)& PPL (C4) & Lat (s) & Mem (G) & PPL (C4) & Lat (s) & Mem (G)\\
           \midrule
           \midrule
           Baseline & 16 & 7.08 & 26.5 & 13.1 & 6.61 & 47.0 & 25.2 & 5.98  & OOM & OOM & 5.62 & OOM & OOM  \\
           \midrule
           GPTQ & 3 & 7.55 & 12.6 & 3.3 & 6.22 & 19.1 & 6.0 & 5.76 & 36.8 & 13.8 & 5.58 & 60.2 & 26.2 \\
            \hc \OURS & 3.02 & 6.32 & 13.6 & 3.4 & 5.60 & 21.2 & 6.1 & 4.66 & 37.8 & 16.1
            & 4.05 &  66.9 & 29.9 \\
            \midrule
           GPTQ (g128) & 3.25 & 6.27 & 110.7 & 3.4 & 5.47 & 176.1 & 6.2 & 4.83 & 500.8 & 14.3 & 4.55 & 955.2 & 27.3\\
           \hc \OURS (0.45\%) & 3.24 & 6.13 & 14.6 & 3.6 & 5.45 & 22.2 & 6.5 & 4.44 & 42.5 & 17.4 & 3.88 & 82.35 & 32.4\\
        \bottomrule
\end{tabular}
}
       
    \end{subtable}

     }
     }
     }
\end{table*}

In Tab.~\ref{tab:hardware-deployment-seqlen1024}, we provide additional hardware profiling results using a sequence length of 1024.
All the experimental setups and details are identical to Sec.~\ref{subsec:deployment} and Tab.~\ref{tab:hardware-deployment-seqlen128}.

Additionally, in Tab.~\ref{tab:hardware-a100}, we demonstrate that our custom CUDA kernels (both including and without including outliers) attain significant speedups of 1.5-2.5$\times$ relative to the fp16 baseline. These results were obtained without any additional optimizations or tuning specifically for the A100, demonstrating how our kernels are easily portable across different GPUs and do not introduce complexity.

\begin{table*}[t!]
\caption{
Matrix-vector kernel runtime (in seconds) for generating 128 tokens, benchmarked on an A100 GPU. 
Our kernel implementation attains 1.5-2.5$\times$ performance speedups relative to the fp16 matrix-vector multiply kernel across different model sizes without any additional optimizations or tuning.
We include GPTQ (with group size 128) without reordering for comparison against the latency of uniform quantization with grouping.
}

\label{tab:hardware-a100}
\centering{
\footnotesize{
\setlength{\tabcolsep}{9pt}{
    \hspace{-8mm}
    
    \begin{subtable}
        \centering
        \scriptsize{
        \vspace{-1mm}
        \begin{tabular}{c|c|ccc}
        \toprule
           \multirow{2}{*}{\textbf{Method}} & \multirow{2}{*}{\textbf{Bit Width}} &  \multicolumn{3}{c}{\textbf{Model}}  \\
           && 7B & 13B & 30B\\
           \midrule
           \midrule
           Baseline & 16 & 1.21 & 2.32 & 5.56 \\
           \midrule
           GPTQ (g128) & 4 & 0.92 & 1.51 & 3.24 \\
           \hb \OURS & 4 & 0.83 & 1.52 & 3.66\\
           \hc \OURS (0.45\%) & 4 & 1.09 & 1.87 & 4.25\\
           \midrule
           GPTQ (g128) & 3 & 0.62 & 1.03 & 2.39 \\
           \hb \OURS & 3 & 0.56 & 0.97 & 2.26\\
           \hc \OURS (0.45\%) & 3 & 0.83 & 1.32 & 2.86\\
        \bottomrule
\end{tabular}
}
       
    \end{subtable}

     }
     }
     }
\end{table*}

\section{Additional Experiment Results}

\subsection{Perplexity Evaluation}
\label{appendix:additional-results}

\begin{table*}[t!]
\label{tab:main}
\caption{Perplexity comparison of LLaMA-30B and 65B models quantized into 4 and 3 bits using different methods including RTN, GPTQ, AWQ and SpQR on C4 and WikiText-2.
We compare the performance of GPTQ, AWQ, and \OURS in groups based on similar model sizes. 
In the first group, we compare dense-only \OURS with non-grouped GPTQ. 
In the subsequent groups, we compare \OURS with different levels of sparsity to GPTQ and AWQ with different group sizes. 
}
\label{tab:llama-7b-main-full}
\vspace{-3mm}
\centering{
\footnotesize{
\setlength{\tabcolsep}{2.7pt}{
    \vspace{3mm}

        \vspace{3mm}
         \begin{subtable}
        \centering
        \scriptsize{
        \vspace{-1mm}
        \begin{tabular}{c|c|cc|c|cc}
        \toprule
            {\textbf{LLaMA-30B}} & \multicolumn{3}{c|}{\textbf{3-bit}}  & \multicolumn{3}{c}{\textbf{4-bit}} \\
            \midrule
           \multirow{2}{*}{\textbf{Method}} & \textbf{Avg. Bits} & \multicolumn{2}{c|}{\textbf{PPL} (↓)} & \textbf{Avg. Bits} & \multicolumn{2}{c}{\textbf{PPL} (↓)} \\
           &  (comp. rate) & C4 & Wiki  &  (comp. rate) & C4 & Wiki\\
           \midrule
           \midrule
           Baseline & 16 & 5.98 & 4.10    & 16 & 5.98 &	4.10 \\
           \midrule
           RTN
               & 3 (5.33) & 28.53 & 14.89
                & 4 (4.00) & 6.33 & 4.54\\
           GPTQ 
                & 3 (5.33) & 7.31 & 5.76 
                & 4 (4.00) & 6.20 & 4.43 \\
            SpQR & - & - & - & 3.89 (4.11) & 6.08 & 4.25 \\
           \hc \OURS
                     & 3.02 (5.31) & \textbf{6.37} &\textbf{4.66} 
                      &  4.03 (3.97) & \textbf{6.06} &\textbf{4.22} \\
           \midrule
           GPTQ (g128) 
                       & 3.25 (4.92)  & 6.47 & 4.83 
                       & 4.25 (3.77)  & 6.07 & 4.24\\
           AWQ (g128) 
                      & 3.25 (4.92)  &  6.38 & 4.63 
                      & 4.25 (3.77)  & 6.05 & 4.21 \\
           \hc \OURS (0.45\%)
                             & 3.25 (4.92) & \textbf{6.23} & \textbf{4.44}
                             &  4.25 (3.77) & \textbf{6.04} & \textbf{4.18} \\

        \bottomrule
        \end{tabular}
     }
     \end{subtable}
         \begin{subtable}
        \centering
        \scriptsize{
        \vspace{-1mm}
        \begin{tabular}{c|c|cc|c|cc}
        \toprule
            {\textbf{LLaMA-65B}} & \multicolumn{3}{c|}{\textbf{3-bit}}  & \multicolumn{3}{c}{\textbf{4-bit}} \\
            \midrule
           \multirow{2}{*}{\textbf{Method}} & \textbf{Avg. Bits} & \multicolumn{2}{c|}{\textbf{PPL} (↓)} & \textbf{Avg. Bits} & \multicolumn{2}{c}{\textbf{PPL} (↓)} \\
           &  (comp. rate) & C4 & Wiki  &  (comp. rate) & C4 & Wiki\\
           \midrule
           \midrule
           Baseline & 16 & 5.62 & 3.53    & 16 & 5.62 & 3.53 \\
           \midrule
           RTN
               & 3 (5.33) & 12.77 &	10.59
                & 4 (4.00) & 5.86 & 3.92\\
           GPTQ 
                & 3 (5.33)  & 6.70 & 5.58 
                & 4 (4.00)  & 5.81	& 4.11\\
            SpQR & 3 (5.33) & - & 4.2$^\dagger$ & 3.90 (4.10) & 5.70 & \textbf{3.68} \\
           \hc \OURS 
                     & 3.02 (5.30) & \textbf{5.99} &\textbf{4.05} 
                     &  4.04 (3.96) & \textbf{5.69} & 3.76 \\
           \midrule
           GPTQ (g128) 
                       & 3.25 (4.92)  & 6.01 & 4.55 
                       & 4.25 (3.77)  & 5.69 & 3.76 \\
           AWQ (g128) 
                      & 3.25 (4.92)  & 5.94 & 4.00
                      & 4.25 (3.77)  & 5.68 & 3.67\\
           \hc \OURS (0.45\%)
                             & 3.24 (4.94) & \textbf{5.84} & \textbf{3.88}
                             &  4.26 (3.76) & \textbf{5.67} & \textbf{3.63} \\

        \bottomrule
        \end{tabular}
        }
     \end{subtable}

     \label{tab:main}
     }
     }
     }
\end{table*}

\begin{table*}[t!]
\caption{Perplexity comparison of LLaMA2 models quantized into 4 and 3 bits using different methods including RTN, GPTQ, AWQ and SpQR on C4 and WikiText-2.
We compare the performance of GPTQ, AWQ, and \OURS in groups based on similar model sizes. 
In the first group, we compare dense-only \OURS with non-grouped GPTQ. 
In the subsequent groups, we compare \OURS with different levels of sparsity to GPTQ and AWQ with different group sizes. 
Note that all GPTQ results are with activation~reordering.
}
\label{tab:llama2}
\vspace{-3mm}
\centering{
\footnotesize{
\setlength{\tabcolsep}{2.7pt}{
    \vspace{3mm}

        \vspace{3mm}
         \begin{subtable}
        \centering
        \scriptsize{
        \vspace{-1mm}
        \begin{tabular}{c|c|cc|c|cc}
        \toprule
            {\textbf{LLaMA2-7B}} & \multicolumn{3}{c|}{\textbf{3-bit}}  & \multicolumn{3}{c}{\textbf{4-bit}} \\
            \midrule
           \multirow{2}{*}{\textbf{Method}} & \textbf{Avg. Bits} & \multicolumn{2}{c|}{\textbf{PPL} (↓)} & \textbf{Avg. Bits} & \multicolumn{2}{c}{\textbf{PPL} (↓)} \\
           &  (comp. rate) & C4 & Wiki  &  (comp. rate) & C4 & Wiki\\
           \midrule
           \midrule
           Baseline & 16 & 6.97 & 5.47    & 16 & 6.97 & 5.47 \\
           \midrule
           RTN & 3 (5.33) & 404.45 & 542.86
                & 4 (4.00) & 7.72 & 6.12 \\
           GPTQ 
                & 3 (5.33)  & 10.45 & 8.97
                & 4 (4.00)  & 7.42 &	5.90 \\
           \hc \OURS 
                     & 3.02 (5.29) & \textbf{7.72} &\textbf{6.18} 
                     &  4.05 (3.95) & \textbf{7.12} &\textbf{5.62} \\
           \midrule
           GPTQ (g128) 
                       & 3.24 (4.93)  & 7.97 & 6.25 
                       & 4.24 (3.77)  & 7.23 & 5.72 \\
           AWQ (g128)
                      & 3.24 (4.93)  & 7.84 & 6.24
                       & 4.24 (3.77)  &7.13 & 5.72\\
           \hc \OURS (0.45\%)
                             & 3.24 (4.93) & \textbf{7.51} & \textbf{5.96}
                             &  4.27 (3.75) & \textbf{7.08} & \textbf{5.57} \\

        \bottomrule
        \end{tabular}
     }
     \end{subtable}
         \begin{subtable}
        \centering
        \scriptsize{
        \vspace{-1mm}
        \begin{tabular}{c|c|cc|c|cc}
        \toprule
            {\textbf{LLaMA2-13B}} & \multicolumn{3}{c|}{\textbf{3-bit}}  & \multicolumn{3}{c}{\textbf{4-bit}} \\
            \midrule
           \multirow{2}{*}{\textbf{Method}} & \textbf{Avg. Bits} & \multicolumn{2}{c|}{\textbf{PPL} (↓)} & \textbf{Avg. Bits} & \multicolumn{2}{c}{\textbf{PPL} (↓)} \\
           &  (comp. rate) & C4 & Wiki  &  (comp. rate) & C4 & Wiki\\
           \midrule
           \midrule
           Baseline & 16 & 6.47 & 4.88    & 16 & 6.47 & 4.88 \\
           \midrule
           RTN 
               & 3 (5.33) & 12.50 & 10.68
               & 4 (4.00) & 6.83 & 5.20\\
           GPTQ
                & 3 (5.33)  & 8.27 & 6.17 
                 & 4 (4.00)  & 6.74 & 5.08 \\
           \hc \OURS
                     & 3.02 (5.30) & \textbf{6.97} &\textbf{5.36}
                      &  4.04 (3.96) & \textbf{6.57} &\textbf{4.99} \\
           \midrule
           GPTQ (g128)
                       & 3.25 (4.92)  & 7.06 & 5.31 
                        & 4.25 (3.77)  & 6.57 & \textbf{4.96}\\
           AWQ (g128) 
                      & 3.25 (4.92)  & 6.94 & 5.32
                      & 4.25 (3.77)  & 6.56 & 4.97\\
           \hc \OURS (0.45\%)
                             & 3.24 (4.94) & \textbf{6.82} & \textbf{5.23}
                             &  4.26 (3.76) & \textbf{6.54} & \textbf{4.96} \\

        \bottomrule
        \end{tabular}
        }
     \end{subtable}
    \vspace{3mm}
     \begin{subtable}
        \centering
        \scriptsize{
        \vspace{-1mm}
        \begin{tabular}{c|c|cc|c|cc}
        \toprule
            {\textbf{LLaMA2-70B}} & \multicolumn{3}{c|}{\textbf{3-bit}}  & \multicolumn{3}{c}{\textbf{4-bit}} \\
            \midrule
           \multirow{2}{*}{\textbf{Method}} & \textbf{Avg. Bits} & \multicolumn{2}{c|}{\textbf{PPL} (↓)} & \textbf{Avg. Bits} & \multicolumn{2}{c}{\textbf{PPL} (↓)} \\
           &  (comp. rate) & C4 & Wiki  &  (comp. rate) & C4 & Wiki\\
           \midrule
           \midrule
           Baseline & 16 & 5.52 & 3.32    & 16 & 5.52 & 3.32 \\
           \midrule
           RTN 
               & 3 (5.33) & 10.02 & 	7.52
               & 4 (4.00) & 5.80	 & 3.67 \\
           GPTQ
                & 3 (5.33)  & 6.69 & 4.86 
                 & 4 (4.00)  & 5.70 & 3.59  \\
           \hc \OURS
                     & 3.02 (5.30) & \textbf{5.83} &\textbf{3.77}
                      &  4.04 (3.96) & \textbf{5.58} &\textbf{3.41} \\
           \midrule
           GPTQ (g128)
                       & 3.25 (4.92)  & 5.87 & 3.88 
                        & 4.25 (3.77)  & 5.59 & 3.42\\
           AWQ (g128) 
                      & 3.25 (4.92)  & 5.81 & 3.74 
                      & 4.25 (3.77)  & 5.58 & 3.41 \\
           \hc \OURS (0.45\%)
                             & 3.24 (4.94) & \textbf{5.73} & \textbf{3.63}
                             &  4.26 (3.76) & \textbf{5.57} & \textbf{3.39} \\

        \bottomrule
        \end{tabular}
        }
     \end{subtable}

     \label{tab:main}
     }
     }
     }
\end{table*}

\begin{table*}[t!]
\caption{Perplexity comparison of OPT models quantized into 4 and 3 bits using different methods including RTN, GPTQ, AWQ and SpQR on C4 and WikiText-2.
We compare the performance of GPTQ, AWQ, and \OURS in groups based on similar model sizes. 
In the first group, we compare dense-only \OURS with non-grouped GPTQ. 
In the subsequent groups, we compare \OURS with different levels of sparsity to GPTQ and AWQ with different group sizes. 
Note that all GPTQ results are with activation reordering.
``div'' means that the perplexity is diverged.
}
\label{tab:opt}
\vspace{-3mm}
\centering{
\footnotesize{
\setlength{\tabcolsep}{2.7pt}{
    \vspace{3mm}

        \vspace{3mm}
         \begin{subtable}
        \centering
        \scriptsize{
        \vspace{-1mm}
        \begin{tabular}{c|c|cc|c|cc}
        \toprule
            {\textbf{OPT-1.3B}} & \multicolumn{3}{c|}{\textbf{3-bit}}  & \multicolumn{3}{c}{\textbf{4-bit}} \\
            \midrule
           \multirow{2}{*}{\textbf{Method}} & \textbf{Avg. Bits} & \multicolumn{2}{c|}{\textbf{PPL} (↓)} & \textbf{Avg. Bits} & \multicolumn{2}{c}{\textbf{PPL} (↓)} \\
           &  (comp. rate) & C4 & Wiki  &  (comp. rate) & C4 & Wiki\\
           \midrule
           \midrule
           Baseline & 16 & 14.72 & 14.62   & 16 & 14.72 & 14.62  \\
           \midrule
           RTN
                      & 3 (5.43)  & div. & div.
                       & 4 (4)  & 24.68 & 48.19\\
           \hc \OURS
                     & 3.04 (5.26) & \textbf{16.42} & \textbf{16.30}
                      &  4.09 (3.91) & \textbf{15.01} &\textbf{14.94} \\
           \midrule
           AWQ (g128)
                      & 3.25 (4.93)  & 16.28 & 16.32
                       & 4.25 (3.77)  & 15.04 & 14.95\\
           \hc \OURS (0.5\%)  
                     & 3.25 (4.92) & \textbf{15.84} & \textbf{15.76} 
                    & 4.30 (3.72) & \textbf{14.94} & \textbf{14.83} \\

        \bottomrule
        \end{tabular}
     }
     \end{subtable}
         \begin{subtable}
        \centering
        \scriptsize{
        \vspace{-1mm}
        \begin{tabular}{c|c|cc|c|cc}
        \toprule
            {\textbf{OPT-2.7B}} & \multicolumn{3}{c|}{\textbf{3-bit}}  & \multicolumn{3}{c}{\textbf{4-bit}} \\
            \midrule
           \multirow{2}{*}{\textbf{Method}} & \textbf{Avg. Bits} & \multicolumn{2}{c|}{\textbf{PPL} (↓)} & \textbf{Avg. Bits} & \multicolumn{2}{c}{\textbf{PPL} (↓)} \\
           &  (comp. rate) & C4 & Wiki  &  (comp. rate) & C4 & Wiki\\
           \midrule
           \midrule
           Baseline & 16 & 13.17 & 12.47   & 16 & 13.17 & 12.47  \\
           \midrule
           RTN 
                  & 3 (5.33)  & div. & div.
                  & 4 (4)  & 17.52 & 16.92\\
           \hc \OURS
                     & 3.04 (5.26) & \textbf{14.45} & \textbf{13.85} 
                      &  4.07 (3.93) & \textbf{13.38} & \textbf{12.80}\\
           \midrule
           AWQ (g128) 
                      & 3.25 (4.93)  & 16.28 & 16.32
                      & 4.25 (3.77)  & 13.39 & 12.73\\
           \hc \OURS (0.5\%) 
                     & 3.25 (4.92) & \textbf{13.88} & \textbf{13.43} 
                     & 4.29 (3.73) & \textbf{13.30} & \textbf{12.60} \\

        \bottomrule
        \end{tabular}
        }
     \end{subtable}

    \vspace{3mm}

 \begin{subtable}
        \centering
        \scriptsize{
        \vspace{-1mm}
        \begin{tabular}{c|c|cc|c|cc}
       \toprule
            {\textbf{OPT-6.7B}} & \multicolumn{3}{c|}{\textbf{3-bit}}  & \multicolumn{3}{c}{\textbf{4-bit}} \\
            \midrule
           \multirow{2}{*}{\textbf{Method}} & \textbf{Avg. Bits} & \multicolumn{2}{c|}{\textbf{PPL} (↓)} & \textbf{Avg. Bits} & \multicolumn{2}{c}{\textbf{PPL} (↓)} \\
           &  (comp. rate) & C4 & Wiki  &  (comp. rate) & C4 & Wiki\\
          \midrule
           \midrule
           Baseline & 16 & 11.74 & 10.86   & 16 & 11.74 & 10.86  \\
           \midrule
           RTN 
                  & 3 (5.33)  & div. & div.
                  & 4 (4) & 13.38 & 12.10 \\
           SpQR  & - & - & -  & 3.94 (4.06) & 11.98 & 11.04\\
           \hc \OURS
                     & 3.02 (5.29) & \textbf{12.44} & \textbf{11.70}
                      &  4.05 (3.96) & \textbf{11.85} & \textbf{11.03}\\
           \midrule
            SpQR & - & - & -  &  4.27 (3.74) & 11.88 & \textbf{10.91} \\
           AWQ (g128) 
                      & 3.25 (4.92)  & 12.30 & 11.41
                      & 4.25 (3.77)  & 11.86 & 10.93\\
           \hc \OURS (0.5\%)
                     & 3.26 (4.90) &\textbf{12.18} & \textbf{11.31}
                      & 4.28 (3.73) & \textbf{11.83} & 10.92 \\

        \bottomrule
        \end{tabular}
     }
     \end{subtable}
         \begin{subtable}
        \centering
        \scriptsize{
        \vspace{-1mm}
        \begin{tabular}{c|c|cc|c|cc}
        \toprule
            {\textbf{OPT-13B}} & \multicolumn{3}{c|}{\textbf{3-bit}}  & \multicolumn{3}{c}{\textbf{4-bit}} \\
            \midrule
           \multirow{2}{*}{\textbf{Method}} & \textbf{Avg. Bits} & \multicolumn{2}{c|}{\textbf{PPL} (↓)} & \textbf{Avg. Bits} & \multicolumn{2}{c}{\textbf{PPL} (↓)} \\
           &  (comp. rate) & C4 & Wiki  &  (comp. rate) & C4 & Wiki\\
           \midrule
           \midrule
           Baseline & 16 & 11.20 & 10.12   & 16  & 11.20 & 10.12   \\
           \midrule
           RTN 
                  & 3 (5.33)  & div. & div.
                  & 4 (4) & 12.35 & 11.32\\
           SpQR  & - & - & - & 3.93 (4.07) & 11.34 & 10.28 \\
           \hc \OURS
                     & 3.02 (5.29) & \textbf{12.69} & \textbf{11.76}
                      &  4.05 (3.96) & \textbf{11.29} & \textbf{10.24}\\
           \midrule
            SpQR  & - & - & - &  4.27 (3.74) & 11.27 & \textbf{10.22}\\
           AWQ (g128) 
                      & 3.25 (4.92)  & 12.61 & 10.67
                      & 4.25 (3.77)  & 11.28 & \textbf{10.22}\\
           \hc \OURS (0.5\%)
                     & 3.26 (4.90) & \textbf{11.57} & \textbf{10.54}
                      & 4.28 (3.73) & \textbf{11.26} & \textbf{10.22} \\

        \bottomrule
        \end{tabular}
        }
     \end{subtable}

    \vspace{3mm}
     
         \begin{subtable}
        \centering
        \scriptsize{
        \vspace{-1mm}
        \begin{tabular}{c|c|cc|c|cc}
        \toprule
            {\textbf{OPT-30B}} & \multicolumn{3}{c|}{\textbf{3-bit}}  & \multicolumn{3}{c}{\textbf{4-bit}} \\
            \midrule
           \multirow{2}{*}{\textbf{Method}} & \textbf{Avg. Bits} & \multicolumn{2}{c|}{\textbf{PPL} (↓)} & \textbf{Avg. Bits} & \multicolumn{2}{c}{\textbf{PPL} (↓)} \\
           &  (comp. rate) & C4 & Wiki  &  (comp. rate) & C4 & Wiki\\
           \midrule
           \midrule
           Baseline & 16 & 10.69 & 9.56   & 16  & 10.69 & 9.56   \\
           \midrule
           RTN
                  & 3 (5.33)  & div. & div.
                   & 4 (4) & 11.90 & 10.98\\ 
           SpQR & - & - & - & 3.94 (4.06) & 10.78 & \textbf{9.54}  \\
           \hc \OURS 
                     & 3.01 (5.31) & \textbf{11.10} & \textbf{10.17}
                     &  4.03 (3.97) & \textbf{10.75}  & 9.65\\
           \midrule
            SpQR & - & - & - &  4.26 (3.76) & 10.73 & \textbf{9.50} \\
           AWQ (g128)
                      & 3.25 (4.92)  & 10.96 & 9.85
                       & 4.25 (3.77)  & 10.75 & 9.59\\
           \hc \OURS (0.5\%)
                     & 3.26 (4.90) & \textbf{10.93} & \textbf{9.77}
                      & 4.28 (3.73) & \textbf{10.72} & 9.61 \\

        \bottomrule
        \end{tabular}
        }
     \end{subtable}

     \label{tab:main}
     }
     }
     }
\end{table*}

In Tab.~\ref{tab:llama-7b-main-full}, we provide the full experimental results on LLaMA~\cite{touvron2023llama}.
Furthermore, in Tab.~\ref{tab:llama2} and \ref{tab:opt}, we provide additional experimental results on LLaMA2~\cite{touvron2023llama2} and OPT~\cite{zhang2022opt} models.
\blfootnote{{$^\dagger$SpQR does not report their near-3-bit performance. However, in the case of 65B model, its 3-bit perplexity on Wikitext-2 can be inferred from the trade-off curve in Figure 8 of their paper. This comparison indicates that the gap between SpQR and \OURS can be larger in the lower-bitwidth regimes.} }

\subsection{5-shot MMLU Evaluation}
\label{appendix:additional-mmlu-5shot}

\begin{table*}[t!]
\caption{Comparison of PTQ methods on five-shot MMLU accuracy applied to Vicuna v1.1 and v1.3.
We add peak memory usage in GB for comparison.
}\label{tab:appendix-mmlu-5shot}
\vspace{1mm}
\centering
\scriptsize{
\setlength{\tabcolsep}{4.5pt}{
\begin{tabular}{c|c|cc|cc|cc|cc|cc}
        \toprule
        \multirow{2}{*}{\textbf{Method}} &  \textbf{Avg.}& \multicolumn{2}{c|}{\textbf{Vicuna-7B (v1.1)}}  & \multicolumn{2}{c|}{\textbf{Vicuna-13B (v1.1)}}  & \multicolumn{2}{c|}{\textbf{Vicuna-7B (v1.3)}}  & \multicolumn{2}{c|}{\textbf{Vicuna-13B (v1.3)}}  & \multicolumn{2}{c}{\textbf{Vicuna-33B (v1.3)}}\\
         & \textbf{bit} & Acc (↑)& Mem (GB, ↓)  & Acc (↑) & Mem (GB, ↓)  & Acc (↑) & Mem (GB, ↓)  & Acc (↑) & Mem (GB, ↓)  & Acc (↑) & Mem (GB, ↓) \\
        \midrule
        \midrule        
        Baseline & 16 & 45.3\% & 12.7 & 50.0\% & 24.6 & 45.6\% & 12.7 & 51.6\% & 24.6 & 60.1\% & OOM \\
        \midrule
        AWQ (g128) & 4.25 & 44.1\% & 3.8 & 48.8\% & 7.2 & 44.8\% & 3.8 & 50.7\% & 7.2 & 59.6\% & 17.2 \\
        \hb \OURS & 4.05 & 44.3\% & 3.8 & 48.4\% & 6.9 & 44.3\% & 3.8 & 50.5\% & 6.9 & 59.6\% & 16.5 \\
        \hc \OURS (0.45\%) & 4.26 & \textbf{44.7\%} & 4.0 & \textbf{49.7\%} & 7.3 & \textbf{44.9\%} & 4.0 & \textbf{51.4\%} & 7.3 & \textbf{60.0\%} & 17.7 \\
        \midrule
        AWQ (g128) & 3.25 & 41.4\% & 3.0 & 46.3\% & 5.7 & 42.5\% & 3.0 & 48.4\% & 5.7 & 56.3\% & 13.3 \\
        \hb \OURS  & 3.02 & 40.4\% & 2.9 & 45.6\% & 5.4 & 41.0\% & 2.9 & 47.4\% & 5.4 & 55.7\% & 12.4 \\
        \hc \OURS (0.45\%)  & 3.24 & \textbf{42.2\%} & 3.1 & \textbf{48.2\%} & 5.8 & \textbf{43.2\%} & 3.1 & \textbf{48.8\%} & 5.8 & \textbf{58.2\%} & 13.7 \\
        \bottomrule
\end{tabular}
}
}
\end{table*}

In Tab.~\ref{tab:appendix-mmlu-5shot}, we provide additional results on 5-shot MMLU  evaluation using the Vicuna v1.1 (7/13B) and Vicuna v1.3 (7/13/33B) models.
We see a similar trend as the zero-shot MMLU evaluation results where \OURS consistently outperforms the baseline quantization methods with the same model size.

\section{Limitations}
 While our empirical results primarily focus on generation tasks, the proposed ideas in this work are not inherently limited to decoder architectures.
 However, we have not yet conducted thorough assessments of our framework's effectiveness on encoder-only or encoder-decoder architectures, as well as other neural network architectures. 
 Additionally, it is important to note that our hardware performance modeling approach relies on a simulation-based method using a roofline model, which entails making simplified assumptions about the hardware's inference pipeline.

\end{document}